\documentclass[pdflatex,sn-mathphys-num]{sn-jnl}

\usepackage{graphicx}%
\usepackage{subcaption}
\usepackage{multirow}%
\usepackage{amsmath,amssymb,amsfonts}%
\usepackage{amsthm}%
\usepackage{mathrsfs}%
\usepackage[title]{appendix}%
\usepackage{xcolor}%
\usepackage{colortbl}
\usepackage{textcomp}%
\usepackage{manyfoot}%
\usepackage{pifont} 
\usepackage{booktabs}%
\usepackage{algorithm}%
\usepackage{algorithmicx}%
\usepackage{algpseudocode}%
\usepackage{listings}%
\usepackage{makecell}
\usepackage{wrapfig}

\definecolor{Gray}{gray}{0.93}

\theoremstyle{thmstyleone}%
\theoremstyle{thmstyletwo}%

\theoremstyle{thmstylethree}%

\begin{document}

\title[Article Title]{Tracking-by-detection in Multi-object Tracking: \\ Survey and Experiments}


\author[1]{\fnm{Yujin} \sur{Yang}}\email{ujin.y@kaist.ac.kr}
\equalcont{These authors contributed equally to this work.}

\author[1]{\fnm{Kyujin} \sur{Shim}}\email{kjshim1028@kaist.ac.kr}
\equalcont{These authors contributed equally to this work.}

\author[1]{\fnm{Kangwook} \sur{Ko}}\email{kw.ko@kaist.ac.kr}

\author*[1]{\fnm{Changick} \sur{Kim}}\email{changick@kaist.ac.kr}

\affil*[1]{\orgdiv{School of Electrical Engineering}, \orgname{Korea Advanced Institute of Science and Technology (KAIST)}, \orgaddress{\city{Daejeon}, \postcode{34141}, \country{South Korea}}}




\abstract{Multi-object tracking (MOT) is an essential computer vision task that simultaneously tracks multiple objects in video sequences, with various applications in surveillance, autonomous navigation, and human-computer interaction. The tracking-by-detection (TBD) paradigm, which combines object detection with temporal association, has emerged as a leading approach, driven by innovative algorithms. Despite recent progress, fair evaluation of TBD-based methods remains a challenge. Many studies introduce modules such as similarity metrics, data association strategies, or motion models, but they are often evaluated under inconsistent protocols, with different baseline trackers, hyperparameters, and datasets. Such inconsistencies obscure the genuine contribution of each module and hinder objective comparison. This survey systematically reviews TBD-based MOT techniques, including similarity measurements, data association, camera motion compensation, and interpolation strategies. Starting from a minimal baseline tracker, we fairly evaluate the contributions of each method across diverse datasets and accumulate well-balanced methods. Our findings establish a strong baseline tracker and provide a foundation for the principled design of robust and versatile MOT systems suitable for real-world deployment.}




\maketitle

\section{Introduction}\label{sec1}

Multi-object tracking (MOT) is a fundamental task in computer vision, focused on simultaneously tracking multiple objects across sequential data, such as video streams. This task is essential for various applications, including intelligent surveillance \cite{features, surv1, sparsetrack, hassan2024multi}, autonomous vehicle navigation \cite{auto_drive1, auto_drive2, auto_drive3, yoltrack}, and human-computer interaction \cite{guidolin2023hi, wu2023referring}. Despite its importance, MOT remains challenging due to persistent issues such as occlusions, significant variations in object appearance, and dynamic motion of both cameras and targets. To address these complexities, numerous approaches have been developed, with the tracking-by-detection (TBD) paradigm emerging as a dominant framework. In TBD, objects are first detected in each frame using advanced object detection algorithms \cite{yolox}, followed by temporal association of these detection results to construct complete object trajectories. By leveraging recent advancements in object detection, feature extraction, and data association, TBD-based trackers \cite{sort, deepsort, bytetrack, bot_sort, tracktrack, sparsetrack, mot-decoupling} consistently achieve state-of-the-art performance, significantly advancing the field of MOT.

Within the TBD framework, each tracker introduces distinctive innovations for MOT. 
SORT \cite{sort} is a seminal tracking-by-detection baseline that demonstrates how a carefully designed yet minimalist pipeline can achieve competitive performance. It employs a linear Kalman filter \cite{kf} with a constant-velocity motion model, where the state vector represents the bounding box center coordinates, scale, aspect ratio, and their respective velocities. For data association, SORT constructs an IoU-based cost matrix between predicted tracklets and current detections, and solves the assignment problem using the Hungarian algorithm \cite{hungarian}. Notably, it relies solely on motion and geometric cues without incorporating appearance features, enabling high computational efficiency and real-time operation.

Building upon this foundation, subsequent trackers introduce additional components to enhance robustness. ByteTrack \cite{bytetrack} improves tracking robustness by associating all detected bounding boxes, including those with low confidence scores, thereby minimizing missed tracks. BoT-SORT \cite{bot_sort} further enhances performance by integrating re-identification models, optimizing the state vector of the Kalman filter, and incorporating camera motion compensation to enhance stability in dynamic scenes. Hybrid-SORT \cite{hybrid_sort} leverages weak cues, such as tracklet confidence and height-modulated Intersection-over-Union (IoU), to improve data association in challenging scenarios with significant object overlap. These advancements highlight the ongoing evolution of TBD-based MOT systems, with novel contributions continually enhancing the reliability and accuracy of object tracking in complex, dynamic environments.

Despite significant advancements in TBD-based MOT methods, evaluating the effectiveness of individual tracking modules remains challenging due to inconsistencies in common components across studies, as shown in Table~\ref{table:comparison}. Variations in detection algorithms, feature extraction methods, and hyperparameter settings hinder direct comparisons and obscure the relative merits of different approaches.
Moreover, many trackers \cite{fairmot, bytetrack, bot_sort, aipt, strong_sort} are primarily evaluated on datasets such as MOT17 \cite{mot16}, which mainly involve linear motion and distinguishable object appearances. This limits the assessment of robustness in more challenging scenarios, such as non-linear motion or visually similar objects, as emphasized in datasets like DanceTrack \cite{dancetrack}.
In addition, commonly used data-splitting protocols construct training and validation sets by temporally dividing the same video sequences, leading to overlap in scene context and object identities. To ensure fair and comprehensive evaluation, standardized protocols with diverse datasets and proper data splits are essential.

\begin{table*}[t!]
    \centering
    \renewcommand{\arraystretch}{1.25}
    \caption{Comparison of common tracking-by-detection components among different trackers. `NMS Thr.' denotes threshold value for Non-Maximum Suppression, `CMC' denotes Camera Motion Compensation, `Post-Proc.' denotes Post-Processing.}
    \resizebox{\linewidth}{!} {
    \begin{tabular}{c|cccccccc}
    \hline
    Tracker                            & Detector                 & Feature Extractor     
                    & NMS Thr.         & CMC & Post-Proc.         & Validation  
                    \\ \hline
    
    SORT            \cite{sort}        & FrCNN    \cite{frcnn}    & -  
                    & N/A              & -   & -                  & MOT15 
                    \\
    DeepSORT        \cite{deepsort}    & FrCNN    \cite{frcnn}    & WRN \cite{wrn} 
                    & N/A              & -   & -                  & MOT16  
                    \\
    ByteTrack       \cite{bytetrack}   & YOLOX-X  \cite{yolox}    & - 
                    & 0.70             & -  & LI \cite{bytetrack} & MOT17, MOT20 
                    \\
    MAATrack        \cite{maatrack}    & CrowdDet \cite{crowddet} & - 
                    & 0.50             & GMC \cite{gmc} & LI      & MOT17, MOT20 
                    \\
    StrongSORT      \cite{strong_sort} & YOLOX-X                  & BoT \cite{bot}     & 0.80            & ECC \cite{ecc}   & GSI + AFLink \cite{strong_sort}                 
                    & \makecell{MOT17, MOT20,\\DanceTrack} 
                    \\
    BoT-SORT        \cite{bot_sort}    & YOLOX-X                  & SBS-50 \cite{fastreid} & 0.65             & GMC & LI                 & MOT17, MOT20                          \\
    Deep OC-SORT    \cite{deep_ocsort} & YOLOX-X                  & SBS-50                                & 0.70             & GMC & LI                 
                    & \makecell{MOT17, MOT20,\\DanceTrack}        
                    \\
    HybridSORT      \cite{hybrid_sort} & YOLOX-X                  & SBS-50 
                    & 0.70             & ECC & LI 
                    & \makecell{MOT17, MOT20,\\DanceTrack} 
                    \\
    BoostTrack      \cite{boosttrack}  & YOLOX-X                  & SBS-50  
                    & 0.3              & ECC & GBI \cite{gbi}     & MOT17, MOT20 
                    \\ \hline
    \end{tabular}}
    \label{table:comparison}
\end{table*}

In this manner, this survey paper systematically reviews and evaluates TBD-based MOT techniques, including key components such as similarity measurements, data association, camera motion compensation, and interpolation strategies.
Our evaluation methodology employs a SORT \cite{sort}-based TBD tracker as an initial baseline and incrementally integrates the most effective method from each functional category step-by-step. We categorize existing techniques according to their role in the tracking pipeline and systematically assess the effectiveness and robustness of each component. To ensure fair comparison, hyperparameters are carefully optimized for each experiment. Through this process, we identify the most effective design choices and progressively incorporate them into a unified baseline tracker.

All experiments are conducted on three diverse MOT benchmarks, namely MOT17 \cite{mot16}, MOT20 \cite{mot20}, and DanceTrack \cite{dancetrack}, which encompass a wide spectrum of motion dynamics, object densities, and scene complexities. 
We construct our splits using entirely different video sequences within each dataset. This protocol mitigates temporal and scene-level overlap between splits, thereby enabling a more rigorous evaluation of generalization across unseen scenarios.

Finally, we establish a generalized baseline tracker that demonstrates consistent performance across all evaluated datasets, providing a solid foundation and practical guidance for the development of robust and versatile MOT systems suitable for real-world deployment.

The main contributions of our article are as follows:
\begin{itemize}
    \item We systematically evaluate the effectiveness of tracking-by-detection (TBD) techniques by incrementally integrating the most effective methods from each functional category into a standardized baseline tracker, enabling precise assessment of individual contributions to MOT performance.

    \item We conduct a fair and comprehensive comparison of recent TBD trackers, and derive practical insights and design guidelines for building reliable and versatile MOT systems in real-world scenarios.

    \item Our study establishes a strong and generalized baseline tracker that performs robustly across diverse datasets, addressing limitations of existing evaluations.
\end{itemize}

The remainder of our paper is formed as follows. Section \ref{sec:back} introduces the problem formulation and SORT-based initial baseline tracker. Then, Section \ref{sec:over} presents the MOT techniques that are compared, and Section \ref{sec:exp_setup} provides an explanation of the object detector, feature extractor, datasets, and metrics we used. Based on this setup, Section \ref{sec:exp} presents the detailed experimental results. Finally, Section \ref{sec:con} summarizes our findings, discusses their implications, and suggests potential directions for future research.


\section{Preliminaries}
\label{sec:back}

\subsection{Problem Formulation}

Following the framework of tracking-by-detection paradigm, we first detect a set of detection results $\mathcal{D}^{t} = \{\textbf{d}^{t}_{1}, ..., \textbf{d}^{t}_{M}\}$ from the current frame at time $t$, where $\textbf{d}^{t}_{i} = (\textbf{b}^{t}_{i}, s^{t}_{i}, \textbf{f}^{t}_{i})$. $\textbf{b}^{t}_{i}$ represents the bounding box location, $s^{t}_{i}$ is the predicted confidence score of the current detection, and $\textbf{f}^{t}_{i}$ indicates the appearance feature extracted from the patch cropped with the corresponding bounding box. As a second step, we predict the current bounding box location of each object track $\mathcal{T}_{j}$ that is tracked until the previous frames by using a motion model. Note that $\mathcal{T}_{j}$ is a set of merged detection results that are determined as the same object. Finally, we properly associate the detection results $\mathcal{D}^{t}$ and object tracks $\mathcal{O} = \{\mathcal{T}_{1}, ..., \mathcal{T}_{N}\}$ through bipartite matching based on their computed pairwise distances, which are measured with their detected or estimated current locations and appearance information. This procedure is iterated over the timeline, thereby resulting in complete tracks from the input video.

\subsection{Baseline Tracker}
\label{sec:baseline}

\begin{algorithm}[t!]
    \caption{Baseline Tracker}
    \label{alg:base_track}
    \footnotesize
    \begin{algorithmic}[1]
        \Require A input video sequence $\mathcal{V}$; Kalman Filter \texttt{KF}; confidence score threshold $\tau_{c}$; matching threshold $\tau_{m}$; minimum length $L_{min}$
        \Ensure Complete tracks $\mathcal{O}$
        \State Initialization: $\mathcal{O} \gets \emptyset$; $\mathcal{O}_{u} \gets \emptyset$;
        \For{frame $f^{t}$ in $\mathcal{V}$}
            \State \# Detect objects in the current frame
            \State $\mathcal{D}^{t} \gets \{\textbf{d}^{t}_{1}, ..., \textbf{d}^{t}_{M} | s_i > \tau_{c}\}$
            \State
            \State \# Predict a current box location for each track
            \State $\mathcal{B}^{t} \gets \emptyset$
            \For{$\mathcal{T}$ in $\mathcal{O}$}
                \State $\mathcal{B}^{t} \gets \mathcal{B}^{t} \cup \{\texttt{KF}(T)\}$
            \EndFor
            \State
            \State \# Distance measurement and association
            \State $C_{1} \gets IoU(\mathcal{D}^{t}, \mathcal{B}^{t})$
            \State Bipartite matching based on $C_{1}$ and $\tau_{m}$
            \State Associate each detection result to the matched track
            \State $\mathcal{O} \gets $ matched tracks
            \State $\mathcal{O}_{u} \gets $ remaining tracks
            \State $\mathcal{D}^{t}_{r} \gets $ remaining detection results
            \State 
            \State \# Predict a current box location for each track
            \State $\mathcal{B}^{t}_{u} \gets \emptyset$
            \For{$\mathcal{T}$ in $\mathcal{O}_{u}$}
                \State $\mathcal{B}^{t}_{u} \gets \mathcal{B}^{t}_{u} \cup \{\texttt{KF}(T)\}$
            \EndFor
            \State 
            \State \# Distance measurement and association
            \State $C_{2} \gets IoU(\mathcal{D}^{t}_{r}, \mathcal{B}^{t}_{u})$
            \State Bipartite matching based on $C_{2}$ and $\tau_{m}$
            \State Associate each detection results to the matched track 
            \State $\mathcal{O}_{u} \gets $ matched unconfirmed tracks
            \State $\mathcal{D}^{t}_{rr} \gets $ remaining detection results
            \State 
            \State \# Confirm the unconfirmed tracks
            \For{$\mathcal{T}$ in $\mathcal{O}_{u}$}
                \If{$len(\mathcal{T}) > L_{min}$}
                    \State $\mathcal{O} \gets \mathcal{O }\cup \{\mathcal{T}\}$
                \EndIf
            \EndFor
            \State $\mathcal{O}_{u} \gets \mathcal{O}_{u} \, \backslash \, \mathcal{O}$
            \State 
            \State \# Initialization
            \For{$\textbf{d}^{t}$ in $\mathcal{D}^{t}_{rr}$}
                \If{$\textbf{d}^{t}.c^{t} > \tau_{c}$}
                    \State $\mathcal{O}_{u} \gets \mathcal{O}_{u} \cup \{\{\textbf{d}^{t}\}\}$
                \EndIf
            \EndFor
        \EndFor
            \State 
        \State Return: $\mathcal{O}$
    \end{algorithmic}
\end{algorithm}

Our initial baseline tracker is based on the SORT \cite{sort} framework, the most basic algorithm of the TBD tracker, and operates by processing each input frame to detect and track multiple objects. Algorithm \ref{alg:base_track} outlines the key steps involved in the tracker, including object detection, track prediction, distance measurement, data association, and track management. In detail, for each frame $f^{t}$ in a video sequence $\mathcal{V}$, we first detect objects resulting in a set of high confidence detection results $\mathcal{D}^t$, whose confidence scores exceed the confidence score threshold $\tau_{c}$ (line 2 to 4). Next, the tracker predicts the current locations of the tracks $\mathcal{O}$ using a Kalman Filter \cite{kf}, and the predicted bounding boxes $\mathcal{B}^{t}$ are used to compute IoU cost matrix $C_{1}$ with the detection results $\mathcal{D}^{t}$ (line 6 to 13). Based on the cost matrix and a matching threshold $\tau_{m}$, bipartite matching is performed with the Hungarian algorithm \cite{hungarian}, and we merge each matched detection result to the corresponding matched track (lines 14 to 15). We also update the set of tracks $\mathcal{O}$ and identify remaining unassociated detection results $\mathcal{D}^{t}_{r}$ (lines 16 to 18).

The tracker then focuses on unconfirmed tracks $\mathcal{O}_{u}$, which are tracked for less than $L_{min}$ number of frames until the last frame. Similar to the previous step, we predict their current locations, measure IoU distances with the remaining detection results $\mathcal{D}^{t}_{r}$, and associate the matched pairs (lines 19 to 28). Then, we update the set of unconfirmed tracks $\mathcal{O}^t_{u}$ only with matched tracks and identify unassociated detection results $\mathcal{D}^t_{rr}$ (lines 29 to 30). These separated association steps prioritize the confirmed tracks to be first associated with the detection results rather than the unconfirmed tracks, which can be short and noisy, blocking their possible disturbance. After the associations, we check the length of each unconfirmed track and move the tracks that exceed a minimum length $L_{min}$ to $\mathcal{O}$ (lines 32 to 38). Finally, remaining unassociated detection results with high-confidence scores are initialized as new tracks and added to $\mathcal{O}_{u}$ (lines 40 to 45). Note that the re-matching of lost tracks is omitted in the pseudocode for simplicity. In the actual implementation, we chase the lost tracks using the motion model and try to match them with the detection results in the first association step for $L_{lost}$ number of frames.


\section{Key Components of Tracking-by-Detection}
\label{sec:over}

The tracking-by-detection (TBD) framework has become a dominant paradigm in modern multi-object tracking, where tracking is formulated as associating detection results across consecutive frames. To improve tracking performance under diverse and challenging scenarios, numerous methods have been proposed, introducing design variations at different stages of the TBD pipeline.

\begin{figure*}[t!]
    \centering
    \includegraphics[width=\textwidth]{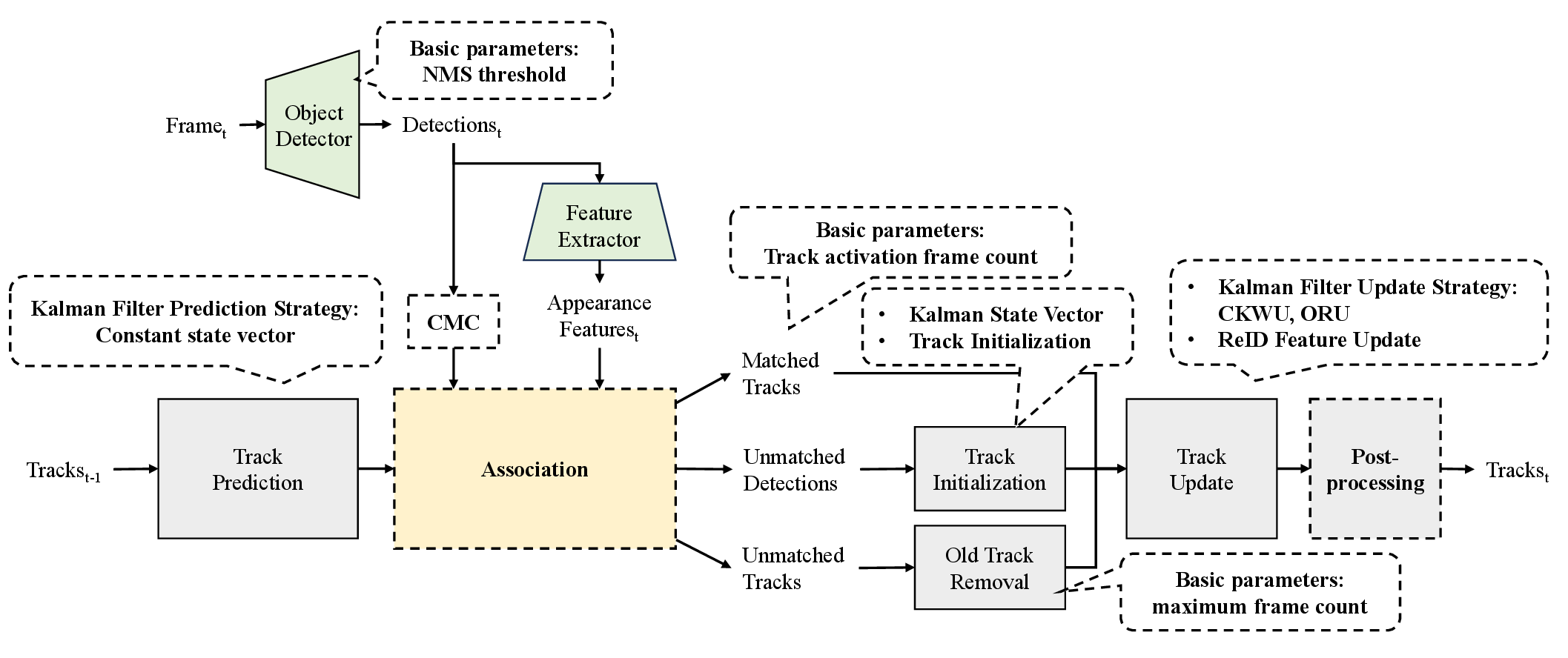}
    \caption{An overview structure of a typical TBD-based MOT tracker. The contents in the dashed boxes are the main techniques that we applied in our experiments.}
    \label{fig:structure}
\end{figure*}

In this section, we categorize these methods based on their functional roles within the tracking process and systematically review representative design choices for each component. Specifically, we focus on key components—such as the Kalman filter state representation, track initialization strategies, data association mechanisms, camera motion compensation, and post-processing techniques—and summarize commonly adopted approaches as well as notable advancements proposed in prior works. As illustrated in Fig.~\ref{fig:structure}, each component corresponds to a specific stage within the overall tracking pipeline.

\subsection{Kalman State Vector}

In MOT, the Kalman filter predicts and updates the states of tracked objects over time. 
The Kalman filter iteratively performs two steps: prediction and update. In the prediction step, the object's state in the next frame is estimated using a motion model, typically assuming constant velocity. In the update step, this prediction is refined by incorporating the associated detection, correcting the estimate based on the observed measurement. 

To represent the current state of a tracking object efficiently, the state vector typically includes parameters describing the object's position, size, and motion.
The design of the Kalman state vector plays a critical role in tracking performance, as it determines which spatial and motion attributes are modeled and propagated over time. 
Commonly used state vectors in MOT include: \begin{itemize}
    \item $(cx, cy, s, a)$: This Kalman state vector is applied for \cite{sort}. Here, $cx$ and $cy$ represent the center coordinates of the object bounding box, $s$ denotes its scale (area), and $a$ represents the aspect ratio. This formulation is useful for tracking objects, such as people or vehicles, where size and shape are essential.
    
    \item $(cx, cy, a, h)$: This Kalman state vector is introduced by \cite{deepsort}. In this variation, $cx$ and $cy$ still represent the center coordinates, $a$ is the aspect ratio, and $h$ is the height of the object. This state vector is effective when the height is a more reliable feature for tracking than the overall scale, particularly in scenes where object height remains consistent.

    \item $(cx, cy, w, h)$: This Kalman state vector is introduced by \cite{bot_sort}. In this configuration, $cx$ and $cy$ define the center of the object, while $w$ and $h$ represent its width and height, respectively. This representation is widely used in object detection tasks, as it directly models the bounding box of the target object.
\end{itemize}


\subsection{Track Initialization}

Track initialization is a crucial step in TBD frameworks, as it determines how new tracks are created from unmatched detection results after association. Traditional TBD methods \cite{bot_sort, bytetrack, biou} rely on heuristic rules to minimize ghost tracks caused by false positives, often applying a higher confidence threshold for initialization than for association. To utilize surrounding information, Occlusion-Aware Initialization (OAI) \cite{improved} enhances the track initialization by incorporating spatial context. OAI calculates the intersection-over-union (IoU) between unmatched detection results and existing track boxes, preventing initialization if the maximum IoU exceeds a predefined threshold, and reducing duplicate tracks. With confidence thresholds and tentative track strategies, OAI offers a robust solution for initializing tracks in complex, densely populated environments.


\subsection{Kalman Filter Prediction \& Update Strategy}

Several strategies have been proposed for updating Kalman state vectors: fixing some parts of the state vectors of inactive tracks \cite{maatrack, conftrack}, Confidence Weighted Kalman Update (CKWU) \cite{conftrack}, and Observation-Centric Re-Update (ORU) \cite{oc_sort}. The first strategy addresses issues caused by occlusion, where changes in box size lead to identity loss; solutions include preserving the height \cite{maatrack} or both height and width of inactive track boxes \cite{conftrack} during prediction. CKWU in ConfTrack \cite{conftrack} adjusts the Kalman filter update by incorporating the confidence score of the detection box. ORU, introduced by OC-SORT \cite{oc_sort}, mitigates accumulated errors in Kalman filter parameters during untracked periods by recalculating Kalman parameters using a virtual trajectory based on observations before and after the untracked period.


\subsection{Basic Parameters}

In MOT, several key parameters are used to enhance tracking accuracy. Non-maximum suppression (NMS) threshold helps eliminate duplicate detection results by filtering out overlapping bounding boxes, keeping only the one with the highest confidence score for each existing object. The minimum frame count for track activation sets the required number of consecutive frames in which an object must be detected to be considered a valid track, ensuring robustness to false positives. Meanwhile, the maximum frame count for lost frames defines how long a track can remain undetected before it is permanently deactivated, allowing for temporary occlusions while avoiding excessive track fragmentation.


\subsection{Camera Motion Compensation}

Camera motion compensation (CMC) is crucial for maintaining tracking accuracy in scenes with significant camera movement. It compensates for the motion of the camera to enhance the stability of object tracking. Two common methods are global motion compensation (GMC) \cite{gmc} and enhanced correlation coefficient maximization (ECC) \cite{ecc}. GMC models global camera motion across the entire scene, offering a simpler but effective correction for consistent background movement, improving overall tracking performance. Conversely, ECC aligns frames by maximizing the correlation between consecutive frames, compensating for complex motion.


\subsection{Main Spatial Distance}

In MOT, spatial distance metrics are critical for assessing the similarity between tracks and detected bounding boxes. Intersection over Union (IoU) usually serves as the standard metric, measuring the overlap between two given boxes, \(A\) and \(B\), as defined by:
\begin{equation} \label{eq:iou}
    IoU = \frac{|A\cap B|}{|A\cup B|}.
\end{equation}

\begin{figure}[t!]
    \centering
    \includegraphics[width=0.5\linewidth]{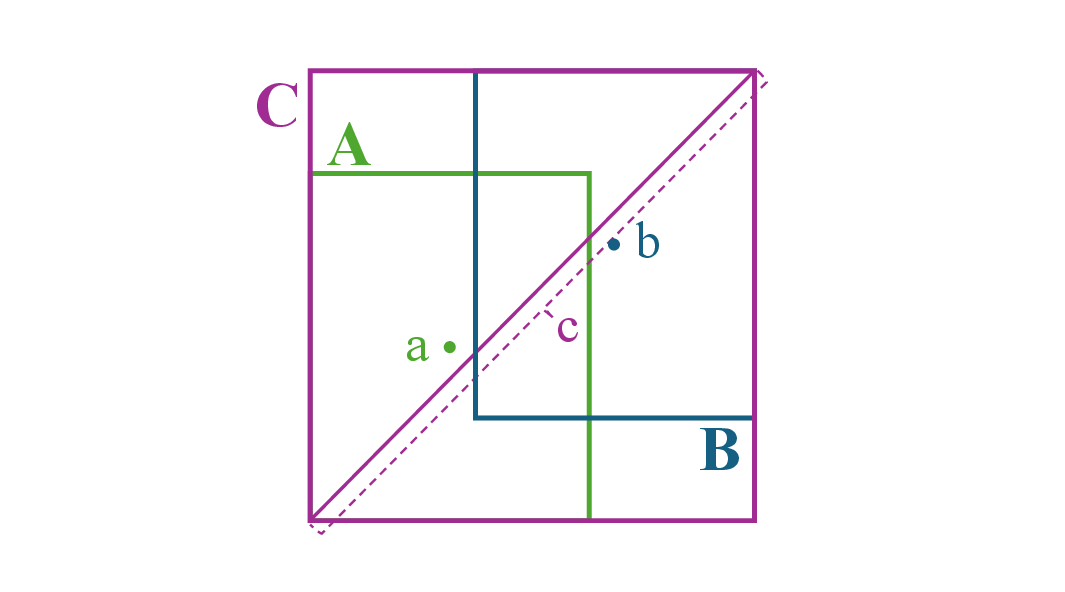}
    \caption{Illustration of bounding boxes. A rectangular \(C\) is the smallest enclosing rectangle that contains boxes \(A\) and \(B\). Dot \(a\) and dot \(b\) are the centers of boxes \(A\) and \(B\), respectively.}
    \label{fig:giou}
\end{figure}

To address limitations in IoU, advanced variants are proposed by adding penalty terms into IoU. Generalized-IoU (GIoU) \cite{giou} considers the smallest enclosing rectangular box \(C\) which contains both boxes A and B as shown in Fig. \ref{fig:giou},
\begin{equation} \label{eq:giou}
    GIoU = IoU - \frac{|C\setminus (A\cup B)|}{|C|}.
\end{equation}
However, GIoU only focuses on the area perspective, Distance-IoU (DIoU) and Complete-IoU (CIoU) \cite{diou} introduce additional penalty terms with the center distance and aspect ratio, respectively, and are defined as
\begin{gather}
    DIoU = IoU - \frac{\rho^2(a, b)}{c^2} \label{eq:diou}, \\
    CIoU = DIoU - \alpha\upsilon \label{eq:ciou},
\end{gather}
where
\begin{gather}
    \upsilon = \frac{4}{\pi^2}(arctan\frac{w^A}{h^A}-arctan\frac{w^B}{h^B})^2 \label{eq:v}, \\
    \alpha = \frac{\upsilon}{(1-IoU)+\upsilon} \label{eq:alpha}.
\end{gather}
Here, as shown in Fig. \ref{fig:giou}, \(a\) and \(b\) are the central points of boxes \(A\) and \(B\), \(c\) is a diagonal length of the enclosing rectangle \(C\), \(\rho(\cdot)\) is the Euclidean distance, and \(w^A\), \(h^A\), \(w^B\), and \(h^B\) represent the width and height of boxes \(A\) and \(B\), respectively. Complete-IoU (CIoU) \cite{diou} factors in both distance and aspect ratio, improving accuracy in object matching. 

\begin{figure}[t!]
    \centering
    \includegraphics[width=0.5\linewidth]{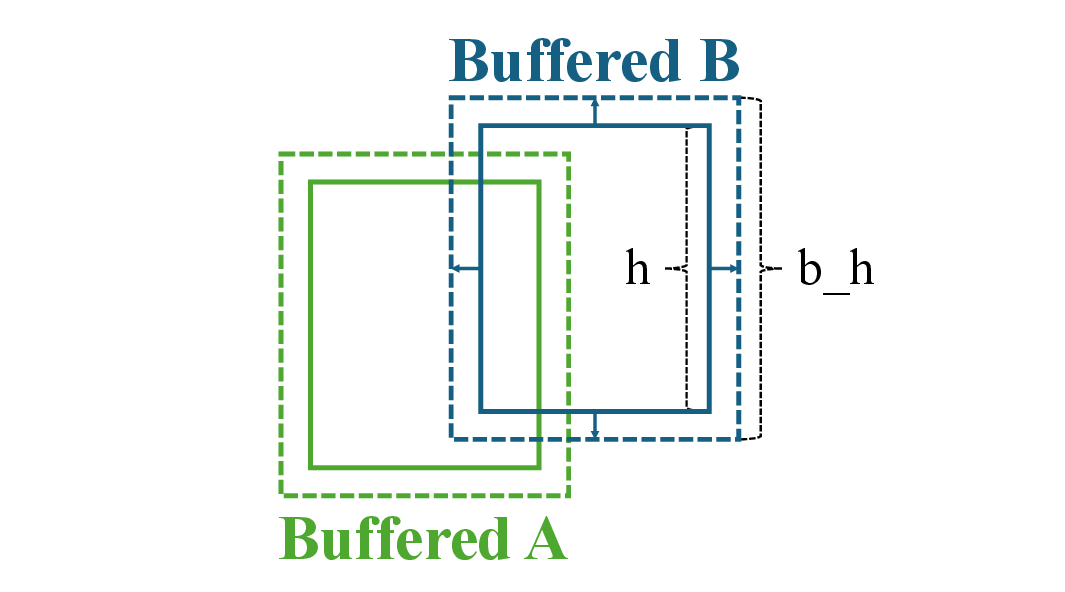}
    \caption{Illustration of buffered bounding boxes. Dashed rectangles are the buffered boxes that are utilized for BIoU\cite{biou}.}
    \label{fig:biou}
\end{figure}

Buffered IoU (BIoU) \cite{biou} further refines IoU by buffering boundaries for better localization, as shown in Fig. \ref{fig:biou}:
\begin{gather}
    BIoU = IoU(\text{buffered } A, \text{buffered } B),\\
    \quad \text{Buffer Scale } b = \frac{b\_h-h}{2h},
\end{gather}
where \(b_h\) is the buffered height and \(h\) is the original height. Similarly, Height Modulated IoU (HMIoU) \cite{hybrid_sort} adjusts IoU by emphasizing height differences, which is particularly effective for tracking objects with distinct vertical size characteristics.
\begin{gather}
    HMIoU = HIoU \cdot IoU,\\
    \quad HIoU = \frac{\text{height of } A\cap B}{\text{height of } C}.
\end{gather}

\subsection{Additional Spatial Distance}

Incorporating advanced spatial distance metrics into IoU calculations can significantly enhance tracking performance. Observation-Centric Momentum (OCM) \cite{oc_sort} mitigates noise in motion direction estimation by leveraging state observations instead of estimations, thereby improving velocity consistency and tracking robustness. Hybrid-SORT \cite{hybrid_sort} refines OCM by integrating a more robust and detailed model for more comprehensive and precise object velocity representation. In addition, it introduces a confidence cost, which is the absolute difference between track and detection confidence scores. Similarly, BoostTrack \cite{boosttrack} presents similarity matrix boost techniques, including Detection-Tracklet Confidence Similarity Boost, Mahalanobis Distance Similarity Boost, and Shape Similarity Boost, which refine the base similarity matrix by adding specially designed similarity metrics to improve the accuracy and reliability of object matching in complex tracking scenarios.


\subsection{Tricks of Spatial Distance}

Recent tricks in spatial distance metrics enhance the robustness of data association. BoostTrack \cite{boosttrack} employs detection confidence boosting to elevate the confidence scores of true positive low-confidence detection results, termed Detecting Likely Objects (DLO), which often arise due to partial occlusions. Specifically, confidence scores of DLO are boosted using the maximum IoU with existing tracks. Conversely, for Detecting Unlikely Objects (DUO) that have low-confidence scores throughout a video due to persistent occlusions, BoostTrack computes their Mahalanobis distance with existing tracks, finds boxes whose all distances exceed a predefined threshold, and increases their confidence scores by considering them as true positives. ConfTrack \cite{conftrack} introduces a Confidence-Fused Cost Matrix, which integrates IoU-based costs with appearance feature-based costs by multiplying the cost matrix with detection confidence scores, thereby constructing a single unified cost matrix. Similarly, PIA \cite{pia} leverages historical detection data alongside current detection results to compute the cost matrix, effectively incorporating temporal information to improve tracking robustness.


\subsection{Appearance Feature Update}

Appearance feature update strategies play a crucial role by enabling robust identity preservation throughout the timeline under varying conditions. Three prominent approaches, namely Feature Bank \cite{deepsort}, Exponential Moving Average (EMA) \cite{ema}, and Dynamic Appearance \cite{deep_ocsort}, offer distinct mechanisms for managing the appearance features of tracked objects. The feature bank \cite{deepsort} approach maintains a static repository of appearance features for each object. Then, it uses them to calculate the maximum similarity value between each detection result and track, providing a stable reference for matching. On the other hand, EMA \cite{ema} is an adaptive mechanism for updating appearance features by blending newly observed appearance features with historical data using a weighted average. This strategy allows the tracker to gradually adapt to changes in object appearance while preserving the earlier observations. More specifically, the update process is defined as,
\begin{equation} \label{eq:ema}
    e^k_i = \alpha e^{k-1}_i + (1-\alpha)f^k_i,
\end{equation}
where \(e^k_i\) is an appearance feature for the \(i^{th}\)  tracks at frame \(k\), and \(f^k_i\) is an appearance embedding of the current detection result. In the following experiments, we set \(\alpha=\)0.9 similar to the previous works \cite{ema, bot_sort}.

The Dynamic Appearance \cite{deep_ocsort} is a confidence-based EMA, where the weight for a feature from each detection result is calculated based on the detection confidence scores. It allows the tracker to concentrate on the features of high-confidence detection. From eq (\ref{eq:ema}), it sets the \(\alpha\) to be a variable \(\alpha_t\), which depends on the confidence score \(s_{det}\) of the current detection result as,
\begin{equation} \label{da}
    \alpha_t = \alpha_f + (1-\alpha_f)(1-\frac{s_{det}-\sigma}{1-\sigma}),
\end{equation}
where \(\sigma\) is a threshold value. In the experiments, we set \(\alpha_f=\) 0.95 following the previous method \cite{deep_ocsort}.


\subsection{Summation}

There have been many trials to achieve a well-balanced summation between the spatial distance matrix \(D_s\) and the appearance distance matrix \(D_a\). The most common approach is a weighted sum using the optimal weight between two distances determined by a grid search as, 
\begin{equation}
    D = \lambda D_a + (1-\lambda) D_s.
\end{equation}
Some trackers, such as ImprAsso \cite{improved}, set the IoU threshold to this weighted sum to prevent wrong matches with objects showing a similar appearance and locating around a current tracking target as,
\begin{equation} \label{eq:sum_thresh}
    D_{i,j} = \begin{cases}
    \lambda d^a_{i,j} + (1-\lambda) d^s_{i,j} &\text{if IoU} > o_{min}\\
    d_{max} + \epsilon &\text{otherwise}.\end{cases}
\end{equation}

Geometric mean \cite{focus} and adaptive weighting \cite{deep_ocsort}, which depends on the discriminativeness of box and track pairs based on the first and second-highest distance, are also suggested as,
\begin{equation}
    D = \sqrt{D_a\cdot D_s},
\end{equation}
\begin{equation}
    D = D_s + [a_w + w]D_a,\\
\end{equation}
respectively, where
\begin{gather}
    w_{i,j} = [z^\text{track}_\text{diff}(D_a, i)+z^\text{det}_\text{diff}(D_a, j)]/2,\\
    z^\text{det}_\text{diff}(D_a, j) = \text{min}(\underset{n}{\mathrm{max}}D_a[n,j]-\underset{m\neq n}{\mathrm{max}}D_a[m,j], \epsilon),\\
    z^\text{track}_\text{diff}(D_a, i) = \text{min}(\underset{n}{\mathrm{max}}D_a[i,n]-\underset{m\neq n}{\mathrm{max}}D_a[i,m], \epsilon).
\end{gather}
Here, \(\epsilon\) is a hyper-parameter to prevent divergence of \(w_{i,j}\) value and set to 0.5, and \(\alpha_w\) is set to 0.75, in our experiments following the original work \cite{deep_ocsort}.

Finally, BOT-SORT \cite{bot_sort} chooses the minimum element among spatial distance and appearance distance as,
\begin{equation} \label{eq:sum_min}
    D_{i,j} = \text{min}\{d^a_{i,j}\text{, }d^s_{i,j}\}.
\end{equation}

\begin{figure*}[t!]
    \centering
    \includegraphics[width=\textwidth]{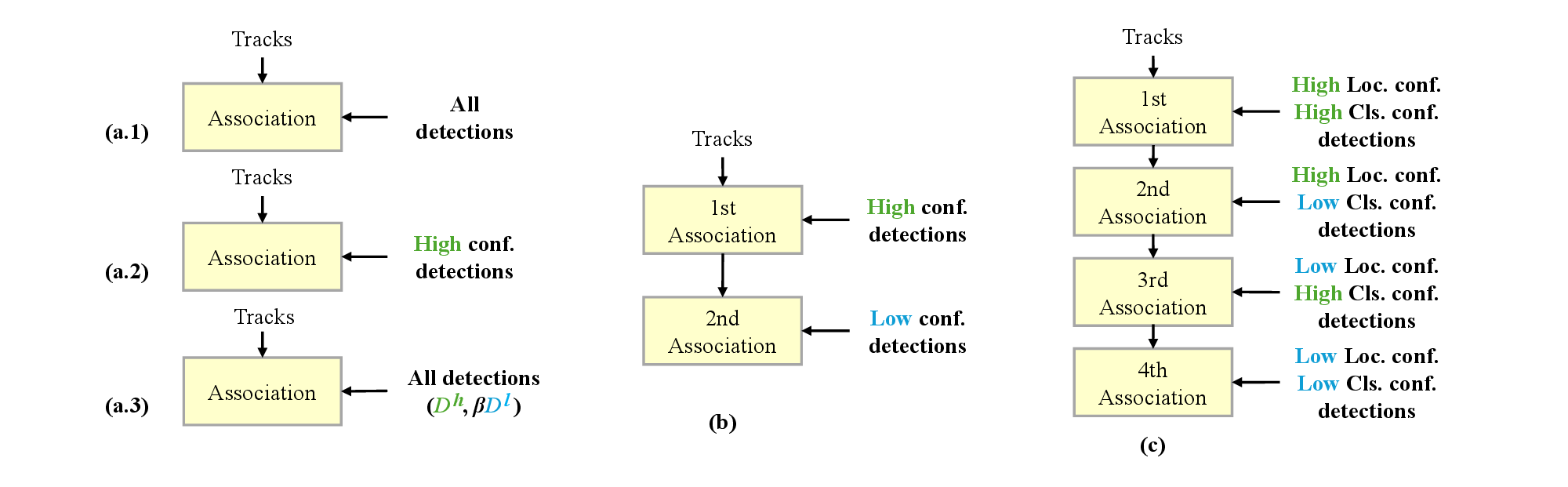}
    \caption{Simplified structures of each association strategy. (a.1) is a basic structure of a one-stage association that uses all detections to match. (a.2) is an advanced one-stage association from BoostTrack \cite{boosttrack} and (a.3) is another advanced one-stage association from Combined Matching \cite{improved}. (b) is a typical two-stage association structure. (c) is a four-stage association structure of LGTrack \cite{lg_track}.}
    \label{fig:associations}
\end{figure*}


\subsection{Association}

Object association is a critical step in MOT, enabling the consistent matching of detected objects to existing tracks to maintain consistent identities across frames. Various association strategies, including one-stage, two-stage, and four-stage approaches, have been developed to balance the trade-off between efficiency, accuracy, and robustness in diverse tracking scenarios. The simplified structures of each association strategy are shown in Fig. \ref{fig:associations}. The one-stage association, proposed in SORT \cite{sort}, matches all existing tracks with new detection results in a single step. Advanced one-stage methods, such as BoostTrack \cite{boosttrack} and Combined Matching \cite{improved}, enhance this process. BoostTrack filters out low-confidence detection results, prioritizing associations between high-confidence detection results and existing tracks. Combined Matching normalizes a cost matrix from low-confidence detection results by applying a scaling factor to align with that of high-confidence detection results, subsequently integrating both matrices to match all detections with tracks at once.

The two-stage association, presented in ByteTrack \cite{bytetrack}, BoT-SORT \cite{bot_sort}, and many other trackers \cite{oc_sort, deep_ocsort}, refines matching through a sequential process. The first stage associates high-confidence detection results with tracks based on spatial proximity or IoU, while the second stage reprocesses unmatched detection results using appearance features or secondary metrics to improve robustness. The four-stage association \cite{lg_track} further extends this framework by systematically addressing distinct tracking challenges across multiple stages. In the first stage, detection results with high localization and classification confidence are matched with existing tracks. The second stage pairs detection results with high localization but low classification confidence with unmatched tracks from the first stage. The third stage associates detection results with low localization but high classification confidence with remaining tracks, and the final stage matches detection results with both low localization and low classification confidence to any remaining tracks, ensuring comprehensive handling of diverse detection scenarios.


\subsection{Score Fusion}

In the final phase of the two-stage association process, new tracks are matched with remaining high-confidence detection results to ensure comprehensive track assignment. Certain approaches, such as those in ByteTrack \cite{bytetrack} and BoT-SORT \cite{bot_sort}, employ a score fusion technique that multiplies detection confidence scores to the IoU cost matrix to prioritize more discriminative detection results as,
\begin{align} 
    C = 1-IoU(D^t, B^t)*s.
\end{align}


\subsection{Post-Processing}

Interpolation is a widely adopted offline post-processing technique in MOT that enhances trajectory continuity and recovers missing detection results caused by occlusions, detection failures, or irregular outputs. Linear Interpolation (LI) \cite{bytetrack}, a basic approach, assumes uniform motion to bridge gaps in trajectories. To incorporate richer motion dynamics, StrongSORT++ \cite{strong_sort} introduces Gaussian-smoothed Interpolation (GSI) \cite{gsi, strong_sort}, while Gradient Boosting Interpolation \cite{gbi, boosttrack} further refines trajectory alignment to follow ground-truth trajectory more properly. These interpolation methods significantly improve the reliability of MOT systems in post-processed data analysis. Additionally, the appearance-free link model (AFLink) \cite{strong_sort} enhances track association by connecting track pairs using spatiotemporal features extracted from their most recent positions and a classifier predicting their association scores to ensure robust track continuity.


\section{Experimental Setup}
\label{sec:exp_setup}

\subsection{Object Detector}

In this work, we use YOLOX \cite{yolox} as an object detector, similar to the previous works \cite{bytetrack, oc_sort, bot_sort, hybrid_sort}. The detector model for each dataset is trained with a combination of the corresponding training set, CrowdHuman \cite{crowdhuman}, and the WiderPerson \cite{widerperson} dataset with eight RTX 3090 GPUs while also following the training configuration of the previous work \cite{bytetrack}. For example, the learning rate starts at 5e-4 and decays with the cosine scheduler, while the total number of training epochs, weight decay, SGD momentum, and batch size are set as 80, 5e-4, 0.9, and 32, respectively. Also, various data augmentation techniques, such as mosaic, random perspective transformation, and mixup, are involved, and the model is initialized with the COCO \cite{coco} pre-trained weight. During the inference stage, thresholds about IoU and confidence scores for non-maximum suppression are set as 0.8 and 0.1, respectively.


\subsection{Feature Extractor}

Following prior works \cite{bot_sort, deep_ocsort, hybrid_sort}, we adopt the SBS-S50 model from the FastReID framework \cite{fastreid} as the feature extractor, leveraging its robust performance for multi-object tracking. The model is fine-tuned on the target dataset using default training configurations initialized from Market-1501 \cite{market1501} pre-trained weights. Specifically, the training is conducted for 60 epochs with a batch size of 64 and learning rate of 3.5e-4 while employing cross-entropy and triplet loss functions \cite{facenet} and a cosine annealing learning rate scheduling. For feature extraction, we first crop patches according to the detection results, resize to 128 × 348 pixels, and apply L2 normalization after the extraction.


\subsection{Datasets}

\begin{figure*}[t!]
    \centering
    \includegraphics[width=\textwidth]{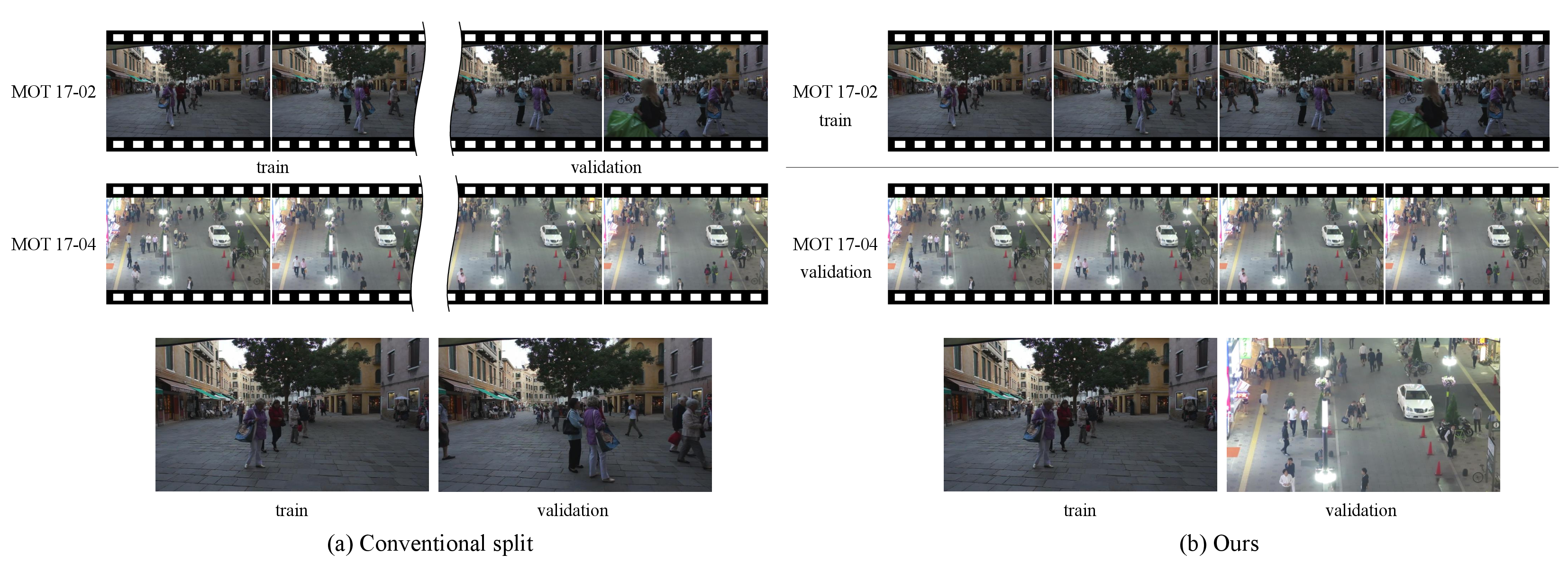}
    \caption{Train and validation split of MOT17: (a) conventional splitting protocol and (b) ours. In the conventional train and validation split used for most previous methods, the first half of each video sequence in the training set is used as the training split, and the remaining halves are used as the validation split. In this scheme, many objects that appear in the training dataset also appear in the validation split, undermining the fundamental purpose of having separate training and validation splits. On the other hand, we split the train set by separating each video sequence, making our evaluation more generalizable.}
    \label{fig:dataset}
\end{figure*}

\begin{table}[t!]
    \centering
    \renewcommand{\arraystretch}{1.25}
    \caption{Characteristics of the three MOT datasets used in the evaluation. ``\# image'', ``\# box'', ``mean \# box'', and ``mean \# overlap'' represent the number of training images, the total number of bounding boxes in the training set, the mean number of bounding boxes per training image, and the mean number of overlapping neighbor bounding boxes per each bounding box, respectively.
    ``object appear.'' denotes the appearances of the objects in the same frame.}
    \begin{tabular}{c|ccc}
    \hline
                    & MOT17       & MOT20       & DanceTrack \\ \hline
    \# image        & 5316        & 8931        & 41796      \\
    \# box          & 112297      & 1134614     & 348930     \\
    mean \# box     & 21.12       & 127.04      & 8.35       \\
    mean \# overlap & 1.81        & 4.26        & 1.81       \\
    camera motion   & O           & X           & O          \\
    object motion   & linear      & linear      & non-linear \\
    mean movement   & 0.036       & 0.022       & 0.045      \\
    object appear.  & distinctive & distinctive & similar    \\ \hline
    \end{tabular}
    \label{table:datasets}
\end{table}

Three prominent and distinctive MOT datasets, MOT17 \cite{mot16}, MOT20 \cite{mot20}, and DanceTrack \cite{dancetrack}, are adopted for every evaluation of our work. MOT17 \cite{mot16} consists of seven training and seven testing sequences, which are filmed in unconstrained environments such as streets and inside malls with diverse camera motion. On the other hand, MOT20 \cite{mot20} includes four videos for each training and test set with highly crowded scenes and heavy occlusion. Finally, DanceTrack \cite{dancetrack} comprises 40, 25, and 35 sequences for training, validation, and testing, respectively. More details about the datasets are shown in Table \ref{table:datasets}. MOT17 and MOT20 are both characterized by linear object motion and distinctive object appearances. However, MOT20 presents problems with higher object density and overlap, and MOT17 offers unstable camera motion. On the other hand, DanceTrack introduces distinctive challenges with its non-linear object motion and similar object appearances combined with complex camera movements. Since the datasets include varying conditions of camera motion, object density, motion patterns, and appearance similarity, each dataset is uniquely valuable for comprehensively assessing the MOT techniques. In addition, it can be a great guide to designing a generalized tracker for real-tracking scenarios by ensuring robust performance across all datasets.

To assess each technique, we construct validation sets for the datasets. However, MOT17 and MOT20 do not provide official validation splits. In most previous works \cite{bytetrack, bot_sort, strong_sort}, a common practice is to divide each training video temporally, using the first half of the frames for training and the latter half for validation, as illustrated in Fig.~\ref{fig:dataset}. 

While this protocol increases the amount of data available for both splits, it introduces substantial overlap in scene context and target objects between training and validation sets. To avoid this issue, we adopt a different splitting strategy: instead of dividing individual videos temporally, \textbf{we separate the dataset at the sequence level and assign different videos to the training and validation sets.} This prevents identity and scene overlap across the splits, enabling a more reliable evaluation of generalization.
More specifically, the sequences ``MOT17-02'', ``MOT17-10'', ``MOT17-11'', and ``MOT17-13'' are used for training, while the remaining sequences ``MOT17-04'', ``MOT17-05'', and ``MOT17-09'' are adopted for validation during our evaluation processes. Similarly, for the MOT20 dataset, the sequences ``MOT20-02'' and ``MOT20-05'' are used for training, the sequences ``MOT20-01'' and ``MOT20-03'' are used for validation.


\subsection{Metrics}

To evaluate various aspects of each technique, we use well-known MOT metrics, including Higher-Order Tracking Accuracy (HOTA) \cite{hota}, Multi-Object Tracking Accuracy (MOTA) \cite{mota}, IDF1\cite{idf1},
Detection Accuracy (DetA)\cite{hota}, and Association Accuracy (AssA)\cite{hota}. 
In this work, we use HOTA as the primary metric because it jointly accounts for detection and association quality (and localization), providing a balanced view of overall tracking performance.


\section{Experiments}
\label{sec:exp}

We conducted a comprehensive evaluation of MOT techniques from Tracking-by-Detection (TBD) approaches, using the baseline tracker detailed in Section~\ref{sec:baseline} as the starting point. We incrementally integrated various techniques, including the Kalman filter, parameter optimization, camera motion compensation, spatial distance metrics, appearance features, summation strategies, association methods, and post-processing techniques. For each sub-task, only the techniques that increase tracking performance in all three datasets are added to the baseline to construct a new baseline for subsequent sub-tasks, ensuring a systematic assessment of the contribution of each component to overall tracking performance.


\subsection{Kalman State Vector}

\begin{table*}[t!]
    \centering
    \renewcommand{\arraystretch}{1.25}
    \caption{Comparison of different Kalman state vectors. The first row is the previous baseline, and the selected option, which will serve as the new baseline tracker for subsequent evaluation, is denoted by a gray row. The increased HOTA scores compared to the previous baseline are underlined, and the best scores are marked in bold.}
    \resizebox{\linewidth}{!}{
        \begin{tabular}{c|ccccc|ccccc|ccccc}
        \hline
                        & \multicolumn{5}{c|}{MOT17} & \multicolumn{5}{c|}{MOT20} & \multicolumn{5}{c}{DanceTrack} \\ \hline
        state vector    & HOTA$\uparrow$ & MOTA$\uparrow$ & IDF1$\uparrow$ 
                        & DetA$\uparrow$ & AssA$\uparrow$
                        & HOTA$\uparrow$ & MOTA$\uparrow$ & IDF1$\uparrow$ 
                        & DetA$\uparrow$ & AssA$\uparrow$
                        & HOTA$\uparrow$ & MOTA$\uparrow$ & IDF1$\uparrow$ 
                        & DetA$\uparrow$ & AssA$\uparrow$ \\ \hline
                        
        (cx, cy, s, a)  & 63.91 & 78.66  & 77.85  & 65.21  & 63.17  
                        & 66.40 & 84.97  & 86.20  & 66.80  & 66.20  
                        & 53.26 & 90.86  & 52.55  & 79.70  & 35.76 \\
                        
        (cx, cy, a, h)  & \underline{64.33} & 79.10  & 78.91  & 65.25  & 63.93  
                        & 66.38             & 84.92  & 87.55  & 66.26  & 66.66  
                        & 47.95             & 89.24  & 51.70  & 71.69  & 32.23 \\
        \rowcolor{Gray} 
        (cx, cy, w, h)  & \underline{\textbf{65.07}} 
                        & \textbf{79.59} & \textbf{79.79} 
                        & \textbf{65.32} & \textbf{65.32}  
                        & \underline{\textbf{67.33}} 
                        & \textbf{85.20} & \textbf{87.91} 
                        & \textbf{67.03} & \textbf{67.82}  
                        & \underline{\textbf{55.71}} 
                        & \textbf{91.47} & \textbf{56.36} 
                        & \textbf{79.90} & \textbf{39.00} \\ \hline
        \end{tabular}}
    \label{table:state}
\end{table*}

\begin{table*}[t!]
    \centering
    \renewcommand{\arraystretch}{1.25}
    \caption{Comparison of different track initialization strategies. Applying OAI increases HOTA across all three datasets.}
    \resizebox{\linewidth}{!}{
        \begin{tabular}{c|ccccc|ccccc|ccccc}
        \hline
                        & \multicolumn{5}{c|}{MOT17} & \multicolumn{5}{c|}{MOT20} & \multicolumn{5}{c}{DanceTrack} \\ \hline
        OAI             & HOTA$\uparrow$ & MOTA$\uparrow$ & IDF1$\uparrow$ 
                        & DetA$\uparrow$ & AssA$\uparrow$
                        & HOTA$\uparrow$ & MOTA$\uparrow$ & IDF1$\uparrow$ 
                        & DetA$\uparrow$ & AssA$\uparrow$
                        & HOTA$\uparrow$ & MOTA$\uparrow$ & IDF1$\uparrow$ 
                        & DetA$\uparrow$ & AssA$\uparrow$ \\ \hline
                        
        \ding{55}       & 65.07  & 79.59  & 79.79  & 65.32  & 65.32   
                        & 67.33  & 85.20  & 87.91  & 67.03  & 67.82 
                        & 55.71  & 91.47  & 56.36  & \textbf{79.90} & 39.00 \\
        \rowcolor{Gray} 
        \ding{51}       & \underline{\textbf{67.00}} 
                        & \textbf{80.24} & \textbf{83.23} 
                        & \textbf{65.90} & \textbf{68.60} 
                        & \underline{\textbf{67.77}} 
                        & \textbf{85.70} & \textbf{88.67} 
                        & \textbf{67.30} & \textbf{68.43} 
                        & \underline{\textbf{57.56}} 
                        & \textbf{91.67} & \textbf{60.04} 
                        & 79.73          & \textbf{41.71} \\ \hline
        \end{tabular}}
    \label{table:init}
\end{table*}

Table \ref{table:state} evaluates the impact of different Kalman state vector representations. The state vectors analyzed include \((cx, cy, s, a)\) \cite{sort}, \((cx, cy, a, h)\) \cite{deepsort}, and \((cx, cy, w, h)\) \cite{bot_sort}. The state vector \((cx, cy, w, h)\) achieves the best performance for all metrics across all datasets. 
These results indicate that incorporating width and height provides the most robust tracking performance, as this representation effectively captures both spatial localization and object dimensions, which are crucial for accurate association. The state vectors \((cx, cy, s, a)\) and \((cx, cy, a, h)\) perform slightly worse.
Especially on DanceTrack, which features similar object appearances and complex motions, the Kalman state with \((cx, cy, w, h)\) outperforms the others, demonstrating its adaptability to challenging scenarios. This result emphasizes the importance of precise dimensional representation in dynamic environments.
The other representations show lower scores in HOTA, emphasizing the importance of precise dimensional representation in dynamic environments.
In summary, the results highlight that the \((cx, cy, w, h)\) state vector offers the best performance in terms of HOTA across all datasets, making it the most effective configuration for accurate and robust tracking. Therefore, the new baseline tracker for the remaining experiments follows the Kalman state vector of \((cx, cy, w, h)\).


\subsection{Track Initialization}

Table \ref{table:init} examines the influence of Occlusion-aware Initialization (OAI) \cite{improved} on tracking performance across the datasets. The results indicate that enabling OAI improves most metrics across all datasets. This highlights the benefits of OAI in dense and complex tracking scenarios, where accurate initialization can prevent identity switches. On the DanceTrack dataset, which presents frequent occlusions, OAI provides the most significant improvement with HOTA increasing from 55.71 to 57.56. This result suggests that OAI is effective in challenging environments by facilitating robust track initialization. Overall, the use of OAI results in consistent gains across most metrics and datasets, with notable improvements. These findings underline the importance of spatial information during track initialization for tracking performance, and OAI is adopted in the following experiments.


\subsection{Kalman Filter Prediction/Update Strategy}

\begin{table*}[t!]
    \centering
    \renewcommand{\arraystretch}{1.25}
    \caption{Comparison of different Kalman filter prediction/update strategies. Trackers with CWKU show improvements across all datasets.}
    \resizebox{\linewidth}{!}{
        \begin{tabular}{c|c|c|ccccc|ccccc|ccccc}
        \hline
                                            \multicolumn{3}{c|}{} & \multicolumn{5}{c|}{MOT17} & \multicolumn{5}{c|}{MOT20} & \multicolumn{5}{c}{DanceTrack} \\ \hline
        Constant    & CWKU      & ORU       & HOTA$\uparrow$  & MOTA$\uparrow$  
                                            & IDF1$\uparrow$  & DetA$\uparrow$  & AssA$\uparrow$
                                            & HOTA$\uparrow$  & MOTA$\uparrow$  & IDF1$\uparrow$  & DetA$\uparrow$  & AssA$\uparrow$
                                            & HOTA$\uparrow$  & MOTA$\uparrow$  & IDF1$\uparrow$  & DetA$\uparrow$  & AssA$\uparrow$ \\ \hline
                                          
        \ding{55}   & \ding{55} & \ding{55} & 67.00 & \textbf{80.24} & 83.23                                                  & 65.90 & 68.60
                                            & 67.77 & 85.70 & 88.67 
                                            & 67.30 & 68.43
                                            & 57.56 & 91.67 & 60.04
                                            & 79.73 & 41.71 \\
                                                
        h           & \ding{55} & \ding{55} & 66.95 & 79.86 & 82.69          
                                            & 65.92 & 68.49
                                            & 67.75 & 85.68 & 88.63 
                                            & 67.30 & 68.40
                                            & \underline{\textbf{58.12}} 
                                            & 91.71 & \textbf{60.53} 
                                            & 79.75 & \textbf{42.52} \\
                                          
        w, h        & \ding{55} & \ding{55} & 66.99 & 79.87 & 83.38          
                                            & 65.80 & 68.69
                                            & 67.77 & 85.71 & 88.67          
                                            & 67.27 & 68.45
                                            & \underline{57.81} 
                                            & 91.65 & 60.10
                                            & 79.91 & 41.96 \\ \hline
        \rowcolor{Gray} 
        \ding{55}   & \ding{51} (0.6)   & \ding{55} & \underline{68.09} 
                                                    & 80.08 & 85.15
                                                    & 66.15 & \textbf{70.53}
                                                    & \underline{67.88} 
                                                    & 85.70 & 88.99
                                                    & 67.36 & 68.60
                                                    & \underline{57.88} 
                                                    & 92.05 & 58.88 
                                                    & 79.88 & 42.11 \\
                                                
        \rowcolor{Gray} 
        \ding{55}   & \ding{51} (0.7)   & \ding{55} & \underline{68.08} 
                                                    & 80.04 & 85.13
                                                    & 66.14 & 70.52
                                                    & \underline{\textbf{68.03}} & 85.75 & \textbf{89.12} 
                                                    & 67.41 & \textbf{68.85}
                                                    & \underline{57.92} 
                                                    & 92.03 & 59.05
                                                    & \textbf{80.01} & 42.12 \\ \hline
                                                
        \ding{55}   & \ding{51} (0.6)   & \ding{51} & \underline{\textbf{68.19}} 
                                                    & 80.07 & \textbf{85.34} 
                                                    & \textbf{66.38} & 70.51
                                                    & \underline{67.99} 
                                                    & \textbf{85.77} & 89.10 
                                                    & \textbf{67.43} & 68.75
                                                    & 57.47 & 91.53 & 59.53 
                                                    & 79.48 & 41.71 \\
                                                
        \ding{55}   & \ding{51} (0.7)   & \ding{51} & \underline{68.11} 
                                                    & 80.00 & 85.20 
                                                    & 66.26 & 70.47
                                                    & \underline{67.98} 
                                                    & 85.75 & 89.00
                                                    & 67.41 & 68.74
                                                    & 57.17 & \textbf{92.07} 
                                                    & 58.17 & 79.99 & 41.03 \\ \hline
        \end{tabular}}
    \label{table:kal_update}
\end{table*}

Table \ref{table:kal_update} provides an analysis of tracking performance based on the inclusion of constant Kalman state vectors \cite{maatrack, conftrack}, Confidence Weighted Kalman Update (CWKU) \cite{conftrack}, and Observation-centric Re-Update (ORU) \cite{oc_sort}. The use of keeping height \cite{maatrack} or both width and height \cite{conftrack} while updating inactive tracks does not produce a significant performance improvement compared to the baseline that updates all states. Although it shows a positive result in DanceTrack, fixing some states delivers slightly lower HOTA and MOTA scores compared to the baseline in MOT17. When CWKU is applied, performance generally improves across most metrics while setting the confidence score threshold to 0.6 or 0.7 following the original work \cite{conftrack}. 
When both CWKU and ORU are applied, the performance shows a mixed trend. For example, the HOTA score increases in MOT17 and MOT20 compared to the tracker only with CKWU. However, the performance on DanceTrack decreases slightly. Overall, the results suggest that fixing the Kalman state vector of inactive tracks does not offer positive effects on the results. In contrast, CWKU enhances tracking performance, while ORU may introduce marginal trade-offs in some cases. In the following experiments, we apply CKWU with a confidence score threshold between 0.6 and 0.7 and choose the best result.


\subsection{Basic Parameters}

\begin{table*}[t!]
    \centering
    \renewcommand{\arraystretch}{1.25}
    \caption{Comparison of different basic parameter settings. The selected option, which will serve as the new baseline tracker for subsequent evaluation, is denoted by a gray row.}
    \resizebox{\linewidth}{!}{
        \begin{tabular}{c|c|c|ccccc|ccccc|ccccc}
        \hline
                                            \multicolumn{3}{c|}{} & \multicolumn{5}{c|}{MOT17} & \multicolumn{5}{c|}{MOT20} & \multicolumn{5}{c}{DanceTrack} \\ \hline
        NMS thresh. & act   & lost max      & HOTA$\uparrow$ & MOTA$\uparrow$ 
                                            & IDF1$\uparrow$ & DetA$\uparrow$ 
                                            & AssA$\uparrow$
                                            & HOTA$\uparrow$ & MOTA$\uparrow$ 
                                            & IDF1$\uparrow$ & DetA$\uparrow$ 
                                            & AssA$\uparrow$
                                            & HOTA$\uparrow$ & MOTA$\uparrow$ 
                                            & IDF1$\uparrow$ & DetA$\uparrow$ 
                                            & AssA$\uparrow$ \\ \hline
                                                 
        0.8         & 3     & frame rate    & 68.09 & 80.08 & 85.15 
                                            & 66.15 & 70.53
                                            & 68.03 & 85.75 & 89.12          
                                            & 67.41 & 68.85
                                            & 57.92 & 92.03 & 59.05 
                                            & 80.01 & \textbf{42.12} \\

        0.7         & 3     & frame rate    & 68.07 & 79.56 & \textbf{85.32} 
                                            & 66.10 & 70.57
                                            & 67.89 & 85.69 & 88.86
                                            & 67.39 & 68.59
                                            & 57.63 & 91.42 & \textbf{59.74} 
                                            & 79.32 & 42.02 \\
        \rowcolor{Gray}
        0.8         & 2     & frame rate    & \underline{\textbf{68.26}} 
                                            & \textbf{80.40} & 85.25          
                                            & \textbf{66.41} & \textbf{70.59}
                                            & \underline{68.15}   
                                            & 85.91 & 89.20          
                                            & 67.54 & 68.96
                                            & \underline{\textbf{57.97}}
                                            & 92.20 & 59.04 
                                            & \textbf{80.15} & 42.11 \\

        0.8         & 2     & 2$\times$frame rate   & \underline{68.20}          
                                            & 80.37 & 85.16          
                                            & 66.40 & 70.50
                                            & \underline{\textbf{68.17}} 
                                            & \textbf{85.94} & \textbf{89.31} 
                                            & \textbf{67.55} & \textbf{69.00}
                                            & 57.80 & \textbf{92.23} 
                                            & 58.79 & 80.05  & 41.91 \\

        0.8         & 2     & 30            & \underline{68.19}          
                                            & 80.37 & 85.12          
                                            & 66.40 & 70.47 
                                            & \underline{\textbf{68.17}} 
                                            & 85.90 & 89.30 
                                            & 67.54 & \textbf{69.00}
                                            & 57.76 & 92.22          
                                            & 58.70 & 80.03  & 41.87 \\ \hline
        \end{tabular}}
    \label{table:param}
\end{table*}

Table \ref{table:param} presents an analysis of the impact of various parameters, specifically the NMS threshold, track activation frame count (act), and the maximum allowable frames for a lost object (lost max). Across all datasets, an NMS threshold of 0.8 consistently yields higher performance than the threshold of 0.7, indicating that a higher NMS threshold generally leads to better tracking results by more effectively suppressing redundant detections.
With an activation frame count of two frames, all aspects of performance improve compared to the configuration with an activation frame count of three, indicating that the tracker can catch new tracks more effectively by setting the activation frame count to two. When adjusting the lost max parameter, it is evident that setting this value to twice the frame rate or 30 frames has minimal impact on performance. For example, in MOT17, the HOTA scores are slightly lower than when we set the lost max parameter as the frame rate of each video. However, longer values overperform in MOT20, indicating longer matching windows are especially suitable for crowded scenes but not for all general scenes. In summary, a higher NMS threshold of 0.8 and reducing the track activation frame count to two frames improves overall tracking performance, and increasing the lost max parameter does not lead to substantial changes in performance. Therefore, we set the NMS threshold as 0.8, the activation count as two frames, and the re-tracking windows as the frame rate length of each video in the new baseline tracker for the following experiments.


\subsection{Camera Motion Compensation}

\begin{table*}[t!]
    \centering
    \renewcommand{\arraystretch}{1.25}
    \caption{Comparison of different CMC methods. Both GMC and ECC bring improvements across all datasets, and GMC makes a bigger gap from the previous baseline tracker.}
    \resizebox{\linewidth}{!}{
        \begin{tabular}{c|ccccc|ccccc|ccccc}
        \hline
                        & \multicolumn{5}{c|}{MOT17} 
                        & \multicolumn{5}{c|}{MOT20} 
                        & \multicolumn{5}{c}{DanceTrack} \\ \hline
        CMC             & HOTA$\uparrow$ & MOTA$\uparrow$ & IDF1$\uparrow$ 
                        & DetA$\uparrow$ & AssA$\uparrow$
                        & HOTA$\uparrow$ & MOTA$\uparrow$ & IDF1$\uparrow$ 
                        & DetA$\uparrow$ & AssA$\uparrow$
                        & HOTA$\uparrow$ & MOTA$\uparrow$ & IDF1$\uparrow$ 
                        & DetA$\uparrow$ & AssA$\uparrow$ \\ \hline
                        
        \ding{55}       & 68.26          & 80.40          & 85.25          
                        & 66.41          & 70.59
                        & 68.15          & 85.91          & 89.20          
                        & 67.54          & 68.96
                        & 57.97          & 92.20          & 59.04          
                        & \textbf{80.15} & 42.11 \\
        \rowcolor{Gray} 
        GMC             & \underline{\textbf{68.61}}      & \textbf{80.89} 
                        & \textbf{85.91} & \textbf{66.51} & \textbf{71.21}
                        & \underline{\textbf{68.26}}      & 85.86
                        & \textbf{89.39} & 67.49          & \textbf{69.22}
                        & \underline{58.23}               & 92.21          
                        & 59.59          & 80.00          & \textbf{42.57} \\
                        
        ECC             & \underline{68.35}               & 80.86          
                        & 85.79          & 66.37          & 70.83
                        & \underline{68.21}               & \textbf{86.01} 
                        & 89.36          & \textbf{67.59} & 69.02
                        & \underline{\textbf{58.27}}      & \textbf{92.27} 
                        & \textbf{59.98} & 80.09          & \textbf{42.57} \\ \hline
        \end{tabular}}
    \label{table:cmc}
\end{table*}

Table \ref{table:cmc} shows the impact of diverse camera motion compensation techniques. The three configurations compared are without CMC, with Global Motion Compensation (GMC) \cite{gmc}, and with Enhanced Correlation Coefficient Maximization (ECC) \cite{ecc}. The results indicate that applying CMC, whether via GMC or ECC, leads to improvements in tracking performance across all datasets, indicating that compensating for camera motion leads to more stable and accurate tracking. 
GMC enhances performance the most in MOT17 and MOT20 with minor improvement in DanceTrack, indicating that compensating for camera motion leads to more stable and accurate tracking. 
Although gains are lower, ECC improves HOTA in MOT17 and MOT20, showing clear benefits in detection and identity preservation. In DanceTrack, ECC slightly outperforms GMC, achieving HOTA of 58.27, compared to 58.23 with GMC.

Overall, the inclusion of CMC, particularly GMC, consistently improves tracking metrics, demonstrating the importance of compensating for camera movement to enhance identity consistency and association accuracy in various tracking scenarios. In the following evaluation, we include GMC in the basic configuration of the baseline tracker.


\subsection{Main Spatial Distance}

\begin{table*}[t!]
    \centering
    \renewcommand{\arraystretch}{1.25}
    \caption{Comparison of matching with different spatial distances. IoU makes the most balanced performance across all datasets.}
    \resizebox{\linewidth}{!}{
        \begin{tabular}{c|ccccc|ccccc|ccccc}
        \hline
                    & \multicolumn{5}{c|}{MOT17} 
                    &\multicolumn{5}{c|}{MOT20} 
                    & \multicolumn{5}{c}{DanceTrack} \\ \hline
        distance    & HOTA$\uparrow$ & MOTA$\uparrow$ & IDF1$\uparrow$ 
                    & DetA$\uparrow$ & AssA$\uparrow$
                    & HOTA$\uparrow$ & MOTA$\uparrow$ & IDF1$\uparrow$ 
                    & DetA$\uparrow$ & AssA$\uparrow$
                    & HOTA$\uparrow$ & MOTA$\uparrow$ & IDF1$\uparrow$ 
                    & DetA$\uparrow$ & AssA$\uparrow$ \\ \hline
        \rowcolor{Gray} 
        IoU         & 68.61 & 80.89  & \textbf{85.91} 
                    & 66.51 & \textbf{71.21}
                    & \textbf{68.26} & 85.86 & \textbf{89.39} 
                    & 67.49 & \textbf{69.22}
                    & 58.23 & 92.21  & 59.59 & 80.00  & 42.57 \\

        Mahalanobis & 63.84 & 78.72  & 78.67 & 65.82  & 62.44
                    & 66.27 & 84.10  & 86.34 & 66.70  & 66.05
                    & 31.77 & 80.21  & 26.29 & 77.46  & 13.08 \\
                        
        GIoU        & 67.78 & 80.57  & 85.04 & 66.54  & 69.51
                    & 68.19 & 85.83  & 89.25 & 67.46  & 69.13
                    & 57.51 & 91.81  & 59.41 & 79.84  & 41.58 \\
                        
        DIoU        & \underline{\textbf{68.64}}      & \textbf{80.97} 
                    & 85.89 & 66.68  & 71.11
                    & 68.13 & 85.59  & 89.17 & 67.32  & 69.14
                    & 57.23 & 92.25  & 58.70 & 80.16  & 41.03 \\
                        
        CIoU        & \underline{\textbf{68.64}}      & \textbf{80.97} 
                    & 85.89 & 66.68  & 71.11
                    & 68.13 & 85.59  & 89.17 & 67.32  & 69.14
                    & 57.24 & 92.25  & 58.70 & 80.14  & 41.06 \\
                        
        BIoU(0.1)   & 68.24 & 80.19  & 85.10 & 66.15  & 70.85
                    & 68.20 & \textbf{86.01} & \textbf{89.39} 
                    & \textbf{67.57} & 69.02
                    & 57.94 & \textbf{92.26} & 59.52  & 80.44 & 41.90 \\
                        
        BIoU(0.2)   & 68.36 & 80.16  & 85.47 & 66.15  & 71.11
                    & 68.16 & 85.93  & 89.32 & 67.48  & 69.04
                    & 57.92 & 92.05  & 59.97 & 79.89  & 42.16 \\
                        
        BIoU(0.3)   & 67.39 & 80.43  & 83.85 & 66.36  & 68.92
                    & 67.97 & 85.83  & 89.04 & 67.39  & 68.74
                    & 57.96 & 91.94  & \textbf{60.21} & 79.42 & 42.48 \\
                        
        HMIoU       & 67.61 & 80.83  & 84.54 & \textbf{66.76} & 68.97
                    & 68.14 & 85.66  & 89.01 & 67.44  & 69.03
                    & \underline{\textbf{58.76}}      & 91.79 
                    & 59.17 & \textbf{80.69} & \textbf{42.97} \\ \hline
        \end{tabular}}
    \label{table:dist_main}
\end{table*}

The evaluation of spatial distance metrics, which are IoU, GIoU~\cite{giou}, DIoU~\cite{diou}, CIoU~\cite{diou}, BIoU~\cite{biou} (with buffer scales of 0.1, 0.2, and 0.3), and HMIoU~\cite{hybrid_sort}, is presented in Table \ref{table:dist_main}. The result reveals IoU as the most robust and effective metric. 
More specifically, IoU consistently outperforms alternatives, achieving strong HOTA scores of 68.61, 68.26, and 58.23 in MOT17, MOT20, and DanceTrack, respectively. 
It suggests that IoU is effective in handling object overlaps and maintaining stable identity tracking across various scenarios. While GIoU, DIoU, and CIoU offer comparable results or marginal improvements in MOT17, they generally underperform in crowded or dynamic settings of MOT20 and DanceTrack. These results suggest that the additional penalty of GIOU for bounding box misalignment beyond the overlap region does not significantly enhance performance. Also, although DIoU and CIoU improve certain spatial constraints, such as box distance, they may not significantly enhance association accuracy in challenging scenarios. BIoU shows competitive performance at lower buffer scales of 0.1 and 0.2 but degrades at higher scales of 0.3. Finally, HMIoU excels in DanceTrack by leveraging height information, yet remains less effective overall. In conclusion, the simplicity and balanced performance of IoU make it the preferred and most efficient choice for general applications, and we also adopt IoU in the following evaluations.


\subsection{Additional Spatial Distance}

\begin{table*}[t!]
    \centering
    \renewcommand{\arraystretch}{1.25}
    \caption{Comparison of matching with different additional spatial distances. ROCM, conf., SMB, and Mahal. denote robust OCM, confidence cost, similarity matrix boost, and Mahalanobis, respectively. The numbers inside the brackets beside the SMB techniques represent the weight values when adding each metric to the base cost matrix. Similarity matrix boost with detection-tracklet confidence similarity draws the most balanced performance.}
    \resizebox{\linewidth}{!}{
        \begin{tabular}{c|c|c|c|ccccc|ccccc|ccccc}
        \hline
        \multicolumn{4}{c|}{} 
        & \multicolumn{5}{c|}{MOT17} 
        & \multicolumn{5}{c|}{MOT20} 
        & \multicolumn{5}{c}{DanceTrack} \\ \hline
        OCM         & ROCM      & conf.     & SMB   
            & HOTA$\uparrow$ & MOTA$\uparrow$ & IDF1$\uparrow$ 
            & DetA$\uparrow$ & AssA$\uparrow$
            & HOTA$\uparrow$ & MOTA$\uparrow$ & IDF1$\uparrow$ 
            & DetA$\uparrow$ & AssA$\uparrow$
            & HOTA$\uparrow$ & MOTA$\uparrow$ & IDF1$\uparrow$ 
            & DetA$\uparrow$ & AssA$\uparrow$ \\ \hline
                                                      
        \ding{55} & \ding{55} & \ding{55} & \ding{55} 
        & 68.61               & 80.89 & 85.91 & 66.51 & 71.21
        & \textbf{68.26}      & 85.86 & 89.39 & 67.49 & \textbf{69.22}
        & 58.23               & 92.21 & 59.59 & 80.00 & 42.57 \\ \hline
                                                      
        \ding{51} & \ding{55} & \ding{55} & \ding{55}                    
        & 68.55               & 80.38 & 85.63 & 66.38 & 71.23
        & 68.12               & 85.94 & 89.16 & 67.53 & 68.90
        & 58.00               & 92.22 & 59.79 & 80.19 & 42.11 \\
                                                      
        \ding{55} & \ding{51} & \ding{55} & \ding{55}                    
        & 68.57               & 80.54 & 85.75 & 66.56 & 71.09
        & 68.17               & 85.87 & 89.26 & 67.47 & 69.07
        & 58.11               & 92.23 & 59.59 & 80.10 & 42.32 \\
                                                      
        \ding{55} & \ding{55} & \ding{51} & \ding{55}                    
        & 68.49               & 80.94 & 86.01  & 66.43 & 71.06
        & 68.20               & 85.97 & 89.29  & \textbf{67.54} & 69.05
        & \underline{58.40}   & \textbf{92.35} & 59.70 & 80.23  & 42.67 \\
                                                      
        \ding{55} & \ding{51} & \ding{51} & \ding{55}                    
        & \underline{68.68}   & 80.64 & 86.25 & 66.55 & 71.34
        & 68.20               & \textbf{85.98} & 89.26 & 67.53 & 69.06
        & \underline{58.25}   & 92.32 & 59.81 & 80.09 & 42.54 \\ \hline
            
        \rowcolor{Gray}
        \ding{55} & \ding{55} & \ding{55} & conf. (0.75)                 
        & \underline{68.62}   & 80.93 & 86.11 & \textbf{66.63} & 71.08
        & \textbf{68.26}      & 85.94 & \textbf{89.46} & 67.51 & 69.21
        & \underline{60.17}   & 92.07 & 62.26 & 80.23 & 45.28 \\
                                                        
        \ding{55} & \ding{55} & \ding{55} & shape (0.75)  
        & \underline{68.76}   & 80.93 & 86.69 & 66.52 & 71.52
        & 68.19               & 85.94 & 89.37 & 67.51 & 69.06
        & \underline{60.32}           & 92.00 & 62.20 & 80.22 & 45.52 \\
        
        \ding{55} & \ding{55} & \ding{55} & Mahal. (0.25)               
        & 68.51               & 80.84 & 85.58 & 66.35 & 71.19
        & 68.17               & 85.64 & 89.27 & 67.36 & 69.17
        & 57.44               & 92.10 & 58.30 & 80.22 & 41.30 \\
        
        \ding{55} & \ding{55} & \ding{55} & conf. (0.50) + shape (0.25)  
        & \underline{\textbf{68.79}}  & \textbf{81.01} 
        & \textbf{86.75}      & 66.57 & \textbf{71.53}
        & 68.24               & 85.94 & 89.40 & 67.51 & 69.16
        & \underline{\textbf{60.37}}  & 92.08 & \textbf{62.71} 
        & 80.26               & \textbf{45.57} \\
        
        \ding{55} & \ding{55} & \ding{55} & conf. (0.50) + Mahal. (0.25) 
        & \underline{68.66}   & 80.81 & 85.94 & 66.46 & 71.38
        & 68.11               & 85.90 & 89.32 & 67.49 & 68.92
        & \underline{58.49}   & 92.31 & 59.12 & \textbf{80.52} & 42.65 \\
        
        \ding{55} & \ding{55} & \ding{55} & shape (0.50) + Mahal. (0.25)
        & 68.07               & 80.65 & 84.64 & 66.38 & 70.27
        & 68.15               & 85.91 & 89.44 & 67.48 & 69.01
        & \underline{58.69}   & 91.76 & 59.72 & 80.05 & 43.18 \\
        
        \ding{55} & \ding{55} & \ding{55} & conf. + shape + Mahal.       
        & 68.53               & 80.86 & 85.92 & 66.47 & 71.11
        & 68.14               & 85.89 & 89.44 & 67.49 & 68.98  
        & \underline{59.79}   & 92.25 & 61.73 & 80.22 & 44.74 \\  \hline
        \end{tabular}}
    \label{table:dist_add}
\end{table*}

Table \ref{table:dist_add} presents the analysis of spatial distance techniques, including Object-Centric Momentum (OCM)~\cite{oc_sort}, robust OCM (ROCM)~\cite{hybrid_sort}, confidence cost (conf.)\cite{hybrid_sort}, and various forms of similarity matrix boosts (confidence, shape, and Mahalanobis distance)\cite{boosttrack}. For the first three methods, OCM, ROCM, and confidence cost, the baseline tracker without these techniques achieves strong HOTA scores in MOT17, MOT20, and DanceTrack. The baseline tracker without these techniques achieves strong HOTA scores of 68.61, 68.26, and 52.83 in MOT17, MOT20, and Dance-
Track, respectively. Introducing OCM slightly reduces HOTA in MOT17 and MOT20, suggesting that basic OCM may not significantly improve the tracking process. When robust OCM is applied with velocity-based adjustments, performance is similar to the original OCM. The application of the confidence cost degrades performance in MOT17 and MOT20 compared to the baseline, although it improves the HOTA score in DanceTrack. 

In the case of the similarity matrix boosts, particularly confidence and shape boosts, deliver the most substantial performance improvements. A confidence boost with a weight of 0.75 elevates HOTA to 68.62 in MOT17 and 60.17 in DanceTrack, with the other metrics. A shape boost also improves HOTA on MOT17 and DanceTrack, but slightly degrades performance on MOT20. Similarly, combining confidence and shape boosts with weights of 0.50 and 0.25 further enhances performance, achieving higher HOTA in MOT17 and DanceTrack. However, performance in MOT20 shows slight declines with these boosts, suggesting sensitivities in crowded scenarios. The combined boost of confidence, shape, and Mahalanobis distance yields marginal gains over individual boosts. The results demonstrate that the most significant performance gains are from applying the confidence similarity matrix boost. Applying this boost technique enhances tracking accuracy, making it especially effective in dynamic scenarios in DanceTrack. Therefore, we include the confidence similarity matrix boost in the new baseline for the succeeding experiments.


\subsection{Tricks of Spatial Distance}

\begin{table*}[t!]
    \centering
    \renewcommand{\arraystretch}{1.25}
    \caption{Comparison of matching with different tricks of spatial distance. The prior baseline without any tricks of spatial distance shows the best performance across all datasets.}
    \resizebox{\linewidth}{!}{
        \begin{tabular}{c|c|c|c|ccccc|ccccc|ccccc}
        \hline
        \multicolumn{4}{c|}{} 
        & \multicolumn{5}{c|}{MOT17} 
        & \multicolumn{5}{c|}{MOT20} 
        & \multicolumn{5}{c}{DanceTrack} \\ \hline
        DLO & DUO & CF & PIA & HOTA$\uparrow$ & MOTA$\uparrow$ & IDF1$\uparrow$ 
                             & DetA$\uparrow$ & AssA$\uparrow$
                             & HOTA$\uparrow$ & MOTA$\uparrow$ & IDF1$\uparrow$ & DetA$\uparrow$  & AssA$\uparrow$
                             & HOTA$\uparrow$ & MOTA$\uparrow$ & IDF1$\uparrow$ & DetA$\uparrow$ & AssA$\uparrow$ \\ \hline
                             
        \rowcolor{Gray} 
        \ding{55}   & \ding{55}      & \ding{55}      & \ding{55} 
        & \textbf{68.62} & 80.93     & 86.11 & \textbf{66.63} & 71.08
        & \textbf{68.26} & 85.94     & 89.46          & 67.51 & \textbf{69.21}
        & \textbf{60.17} & 92.07     & \textbf{62.26} & 80.23 & \textbf{45.28} \\ \hline
        
        \ding{51}   & \ding{55}      & \ding{55}      & \ding{55} 
        & 68.53     & \textbf{80.94} & \textbf{86.17} & 66.45 & \textbf{71.14}
        & 68.23     & \textbf{85.98} & \textbf{89.49} & \textbf{67.54} & 69.12
        & 59.56     & 92.03          & 62.02          & 80.19          & 44.41 \\
        
        \ding{55}   & \ding{51}      & \ding{55}      & \ding{55} 
        & 68.50     & 80.89          & 86.14          & 66.45          & 71.08
        & 68.21     & 85.81          & 89.41          & 67.44          & 69.17
        & 59.84     & \textbf{92.47} & 61.75 & \textbf{80.34}          & 44.74 \\
        
        \ding{51}   & \ding{51}      & \ding{55}      & \ding{55} 
        & 68.37     & 80.91          & 85.88          & 66.29          & 70.97
        & 68.19     & 85.81          & 89.42          & 67.44          & 69.12
        & 59.43     & 92.02          & 61.29          & 80.09          & 44.27 \\ \hline
                                                      
        \ding{55}   & \ding{55}     & \ding{51}       & \ding{55} 
        & 66.88     & 79.45         & 83.81           & 65.57          & 68.67
        & 67.27     & 84.92         & 88.03           & 66.81          & 67.91
        & 60.03     & 91.58         & 61.30           & 80.10          & 45.14 \\ \hline
                                                      
        \ding{55}   & \ding{55}     & \ding{55}       & \ding{51} 
        & 68.09     & 80.25         & 84.98           & 66.08          & 70.62
        & 68.15     & 85.92         & 89.05           & 67.53          & 68.96
        & 59.12     & 91.70         & 59.91           & 80.19          & 43.76 \\ \hline
        \end{tabular}}
    \label{table:dist_trick}
\end{table*}

Table~\ref{table:dist_trick} evaluates the impact of spatial distance enhancement tricks, which are Detecting Likely Objects (DLO)~\cite{boosttrack}, Detecting Unlikely Objects (DUO)~\cite{boosttrack}, Confidence Fused Cost Matrix (CF)~\cite{conftrack}, and Past Information Aggregation (PIA)~\cite{pia}, with the baseline configuration without any tricks. While the baseline achieves HOTA scores of 68.62, 68.26, and 60.17, respectively, applying DLO and DUO alone yields a slight decrease in the HOTA scores. Combining DLO and DUO even more reduces performance, with HOTA dropping to 68.37 in MOT17 and 59.43 in DanceTrack, suggesting potential conflicts in their adjustments. In addition, CF significantly and consistently underperforms compared to the baseline across all datasets, likely due to over-filtering valid detection result. PIA similarly shows performance degradation with HOTA scores across all benchmarks. It suggests that linear assignment with an aggregated cost matrix may result in more mismatches between tracks and detections. 
In summary, these results provide the benefit of few individual DLOs and DUOs, while CFs and PIA degrade tracking accuracy.
Therefore, we maintain the configuration of the current baseline for the following analysis.


\subsection{Appearance Feature Update}

\begin{table*}[t!]
    \centering
    \renewcommand{\arraystretch}{1.25}
    \caption{Comparison of matching with different appearance feature update strategies. Even though Feature Bank with a large number of features $n$ shows great tracking quality, it takes a much longer time to inference so it's hard to employ for real-time tracking. EMA shows the most balanced performance across all datasets.}
    \resizebox{\linewidth}{!}{
        \begin{tabular}{c|c|ccccc|ccccc|ccccc}
        \hline
         \multicolumn{2}{c|}{}
         & \multicolumn{5}{c|}{MOT17} 
         & \multicolumn{5}{c|}{MOT20} 
         & \multicolumn{5}{c}{DanceTrack} \\ \hline
        Feature Update & n  & HOTA$\uparrow$ & MOTA$\uparrow$ & IDF1$\uparrow$ 
                            & DetA$\uparrow$ & AssA$\uparrow$
                            & HOTA$\uparrow$ & MOTA$\uparrow$ & IDF1$\uparrow$ 
                            & DetA$\uparrow$ & AssA$\uparrow$
                            & HOTA$\uparrow$ & MOTA$\uparrow$ & IDF1$\uparrow$ 
                            & DetA$\uparrow$ & AssA$\uparrow$ \\ \hline
                           
        Feature Bank & 100  
            & 67.38             & \textbf{80.49} & 84.19 & 65.63 & 69.68
            & 65.48             & 85.01          & 84.24 & 66.94 & 64.27
            & 34.10             & 69.90          & 26.36 & 72.13 & 16.22 \\
                           
        Feature Bank & 50   
            & 67.74             & 80.39          & 84.69 
                                & 65.77          & 70.27
            & 65.84             & \textbf{85.23} & 84.48 
                                & \textbf{67.19} & 64.74
            & \textbf{34.98}    & 71.09          & \textbf{27.41} 
                                & 71.79          & \textbf{17.13} \\ \hline
                           
        Feature Bank & 10  
            & 67.42             & 80.14          & 84.26 
                                & 65.57          & 69.81
            & \textbf{66.39}    & 85.18 & \textbf{85.58} 
                                & 67.11          & \textbf{65.89}
            & 31.41             & \textbf{72.30} & 23.52 
                                & \textbf{73.23} & 13.56 \\
        \rowcolor{Gray} 
        EMA          & -   
            & \textbf{68.53}    & 79.89          & \textbf{85.47} 
                                & \textbf{66.30} & \textbf{71.29}
            & 64.96             & 84.42          & 82.99          
                                & 66.92          & 63.30
            & 31.59             & 61.51          & 24.19          
                                & 68.58          & 14.62 \\
                           
        DA           & -   
            & 65.64             & 79.07          & 80.43          
                                & 65.94          & 65.85
            & 57.35             & 80.84          & 70.38          
                                & 65.92          & 50.22
            & 26.13             & 53.03          & 18.90          
                                & 66.86          & 10.28 \\ \hline
        \end{tabular}}
    \label{table:feature_update}
\end{table*}

Table~\ref{table:feature_update} evaluates the impact of appearance feature update strategies, notably static Feature Bank \cite{deepsort} with 10, 50, and 100 features, Exponential Moving Average (EMA) \cite{ema}, and Dynamic Appearance (DA) \cite{deep_ocsort}, using the ResNeSt\cite{zhang2022resnest}-based SBS-S50 feature extractor from the FastReID framework \cite{fastreid}. In this experiment, the cost matrices are only based on cosine similarity between detection and track feature vectors for more transparent examination. The static Feature Bank strategy finds that the feature banks with more stored features achieve better performance on complicated scenarios, such as DanceTrack.
For example, a feature bank with 50 features achieves HOTA of 34.98 in DanceTrack. However, its computational inefficiency with more than 10 features, even processing fewer frames than the frame rate in our empirical evaluations, renders it unsuitable for online MOT. 

Among online-MOT-available methods, EMA delivers the best balanced performance. In MOT17, it achieves the best HOTA with 68.53, demonstrating superior tracking stability and identity preservation through smoothed feature updates. Although its performance in MOT20 and DanceTrack indicates limitations in crowded and dynamic scenarios compared to the baseline, it's slightly better than the feature bank with 10 features in DanceTrack. Finally, DA consistently underperforms, with HOTA scores in all datasets, likely due to noise introduced by frequent updates incorporating detection confidence. Thus, EMA emerges as the most effective strategy, balancing robustness and efficiency, while the static Feature Bank requires strong computational overhead, and DA struggles with instability across diverse tracking conditions.


\subsection{Summation}

\begin{table*}[t!]
    \centering
    \renewcommand{\arraystretch}{1.25}
    \caption{Comparison of matching with different summation strategies. The weighted sum of the spatial distance and the appearance distance makes the most balanced improvement.}
    \resizebox{\linewidth}{!}{
        \begin{tabular}{c|ccccc|ccccc|ccccc}
        \hline
        & \multicolumn{5}{c|}{MOT17} 
        & \multicolumn{5}{c|}{MOT20} 
        & \multicolumn{5}{c}{DanceTrack} \\ \hline
        Summation             & HOTA$\uparrow$ & MOTA$\uparrow$ & IDF1$\uparrow$ 
                              & DetA$\uparrow$ & AssA$\uparrow$
                              & HOTA$\uparrow$ & MOTA$\uparrow$ & IDF1$\uparrow$ & DetA$\uparrow$ & AssA$\uparrow$
                              & HOTA$\uparrow$ & MOTA$\uparrow$ & IDF1$\uparrow$ & DetA$\uparrow$ & AssA$\uparrow$ \\ \hline
                              
        only spatial dist.    & 68.62          & 80.91          & 86.19          
                              & 66.62          & 71.12
                              & 68.26          & 85.94          & 89.46          
                              & 67.51          & 69.21
                              & 60.17          & 92.07          & 62.26          
                              & 80.23          & 45.28 \\
                              
        only appearance dist. & 68.53          & 79.89          & 85.47          
                              & 66.30          & 71.29
                              & 64.95          & 84.36          & 83.22          & 66.81          & 63.37
                              & 31.59          & 61.51          & 24.19          & 68.58          & 14.62 \\ \hline
        \rowcolor{Gray} 
        weighted sum          
                & \underline{\textbf{68.94}}   & 80.66          & 86.24          
                & 66.42          & \textbf{72.01}
                & \underline{68.42}            & 85.81          & 89.57          
                & 67.48          & \textbf{69.56}
                & \underline{\textbf{62.85}}   & \textbf{92.09} & \textbf{65.94} & 80.34          & \textbf{49.35} \\

        thresh. weighted sum  
                & \underline{\textbf{68.94}}  & 80.65           & 86.24          & 66.42          & \textbf{72.01}
                & \underline{68.43}           & 85.88           & \textbf{89.59} & 67.54          & 69.51
                & \underline{62.52}           & 92.08           & 65.34          & 80.17          & 48.95 \\  
        
        geometric mean        
                & \underline{68.84}            & 80.69          & 86.06          
                & 66.57          & 71.66
                & \underline{\textbf{68.46}}   & \textbf{85.96} & 89.63          
                & \textbf{67.58} & 69.53
                & \underline{62.29}            & 91.90          & 65.39          & 79.86          & 48.77 \\
                              
        adaptive weighting    
                & \underline{68.67}            & 80.81          & 85.50          & \textbf{66.78} & 71.07
                & 68.25                        & 85.92          & 89.26          & 67.56          & 69.14
                & \underline{61.44}            & 92.03          & 63.08          & \textbf{80.68} & 46.96 \\
                              
        minimum               
                & \underline{68.84}           & \textbf{80.94}  & \textbf{86.33} & 66.41          & 71.83
                & \underline{68.36}           & 85.69           & 89.44          & 67.40          & 69.53
                & 58.15                       & 91.38           & 60.08          & 79.74          & 42.56 \\ \hline
        \end{tabular}}
    \label{table:summation}
\end{table*}

Table~\ref{table:summation} shows the impact of different distance summation strategies, including only spatial distance, only appearance distance, weighted sum, threshold-weighted sum \cite{improved}, geometric mean \cite{focus}, adaptive weighting \cite{deep_ocsort}, and minimum element among spatial distance and appearance distance\cite{bot_sort}. 
Using only spatial distance yields robust performance with HOTA in all benchmarks, while relying solely on appearance distance significantly underperforms, particularly in DanceTrack. This highlights the limitations of relying only on appearance features in dynamic scenes, where spatial positioning provides robust tracking. From this perspective, combining two features is expected to result in better performance.

The weighted sum of spatial distance and appearance distance approaches balance spatial and appearance information, resulting in the highest HOTA scores across all configurations. The weighted sum achieves HOTA of 68.94 in MOT17, 68.42 in MOT20, and 62.85 in DanceTrack. This indicates that integrating spatial and appearance information through weighted calculations provides an optimal tracking performance. The weighted sum method, with an optimized weight \(\lambda=0.7\) determined via grid search.
The geometric mean is slightly below the weighted sum, but shows a noticeable improvement in performance over the baseline. 
The threshold-weighted sum configuration and geometric mean perform comparably to the weighted sum, but this approach does not significantly outperform the standard weighted sum in DanceTrack. 
The adaptive weighting method shows lower scores in all datasets. The minimum-based approach achieves slightly lower scores than the previous baseline, particularly in DanceTrack, indicating that taking the minimum distance may overly limit the amount of information in complex environments. Overall, the weighted sum method achieves the best balance of spatial and appearance cues, consistently yielding high HOTA scores across all datasets. These results suggest that combining spatial and appearance information with balanced weighting is crucial, and we choose the weight summation as our combining method between spatial and appearance distances for the following evaluations.


\subsection{Association}

\begin{table*}[t!]
    \centering
    \renewcommand{\arraystretch}{1.25}
    \caption{Comparison of matching with different association strategies. 2-stage association is the best balanced strategy for TBD. }
    \resizebox{\linewidth}{!}{
        \begin{tabular}{c|ccccc|ccccc|ccccc}
        \hline
        & \multicolumn{5}{c|}{MOT17} 
        & \multicolumn{5}{c|}{MOT20} 
        & \multicolumn{5}{c}{DanceTrack} \\ \hline
        Association             & HOTA$\uparrow$ & MOTA$\uparrow$ 
                                & IDF1$\uparrow$ & DetA$\uparrow$ 
                                & AssA$\uparrow$
                                & HOTA$\uparrow$ & MOTA$\uparrow$ 
                                & IDF1$\uparrow$ & DetA$\uparrow$ 
                                & AssA$\uparrow$
                                & HOTA$\uparrow$ & MOTA$\uparrow$ 
                                & IDF1$\uparrow$ & DetA$\uparrow$ 
                                & AssA$\uparrow$ \\ \hline
                                    
        1-stage                 & 68.94          & 80.66          & 86.24                                         & 66.42          & 72.01
                                & 68.42          & 85.81          & 89.57        
                                & 67.48          & 69.56
                                & 62.85          & 92.09          & 65.94        
                                & 80.34          & 49.35 \\
                                     
        1-stage 
        (unconfirmed tracks)    & 68.88          & 80.67          & 86.09                                         & 66.34          & 71.97          & \underline{68.44} 
                                & 85.81          & 89.62        
                                & 67.48          & 69.59          & \underline{62.98} 
                                & 92.67          & 65.78        
                                & \textbf{80.85} & 49.26 \\
                                     
        1-stage (BoostTrack)    & 67.74          & 79.02          & 84.72                                         & 65.14          & 70.87
                                & 66.85          & 83.44          & 88.08        
                                & 65.87          & 68.03
                                & 61.69          & 90.52          & 63.35        
                                & 79.79          & 47.84 \\
                                     
        1-stage 
        (Combined matching)     & 68.83          & 80.70          & 86.09                                         & 66.48          & 71.73
                                & 68.40          & 85.88          & 89.51                 & 67.55          & 69.45
                                & \underline{\textbf{63.58}}      & 92.07        
                                & \textbf{66.87} & 80.23          & \textbf{50.56} \\
                                
        \rowcolor{Gray} 
        2-stage                 & \underline{\textbf{69.35}}      & \textbf{81.04} 
                                & \textbf{87.16}                  & \textbf{66.77} 
                                & \textbf{72.45}
                                & \underline{\textbf{68.58}}      & \textbf{86.07} 
                                & \textbf{89.93}                  & \textbf{67.65} 
                                & \textbf{69.72}
                                & \underline{63.27}               & \textbf{92.71} 
                                & 66.07                           & 80.84          
                                & 49.71 \\
                                     
        4-stage                 & 67.78         & 79.96           & 84.56          
                                & 65.82         & 70.24
                                & 67.24         & 84.27           & 88.16          
                                & 66.47         & 68.20
                                & 61.39         & 91.05           & 64.36          
                                & 79.27         & 47.74 \\ \hline
        \end{tabular}}
    \label{table:association}
\end{table*}

The evaluation of various association strategies, as presented in Table~\ref{table:association}, demonstrates that the 2-stage association strategy\cite{bytetrack} achieves the highest HOTA scores across all datasets, with 69.35 in MOT17, 68.58 in MOT20, and 63.27 in DanceTrack. 
These results highlight its superior balance of detection and association accuracy by prioritizing high-confidence detection results in the first stage, making it the most effective strategy overall. The 1-stage association process\cite{sort}, which is computationally simpler, yields slightly lower performance, indicating that the additional matching with low-confidence detection results through the second stage enhances the ability to handle complex scenarios and maintain the identity consistency of the tracker. In contrast, the 1-stage strategy from BoostTrack\cite{boosttrack} and the 4-stage strategy\cite{lg_track} perform less favorably, suggesting that using only high-confidence detection results or overcomplicating the association process may introduce noise or overfitting that can diminish robustness. 
The 1-stage combined matching strategy\cite{improved}, which penalizes low-confidence detection results, performs competitively but offers marginal improvements over the basic 1-stage approach except in DanceTrack. These results demonstrate that the 2-stage association strategy provides the best balance of accuracy and robustness. It shows giving chances to low-confidence detections after prioritizing high-confidence detection results can guarantee robust tracking performance. Therefore, the tracker that will serve as the new baseline adopts the 2-stage strategy.


\subsection{Score Fusion}

\begin{table*}[t!]
    \centering
    \renewcommand{\arraystretch}{1.25}
    \caption{Comparison between matching with/without score fusion. Adopting score fusion makes slightly better HOTAs.}
    \resizebox{\linewidth}{!}{
        \begin{tabular}{c|ccccc|ccccc|ccccc}
        \hline
        & \multicolumn{5}{c|}{MOT17} 
        & \multicolumn{5}{c|}{MOT20} 
        & \multicolumn{5}{c}{DanceTrack} \\ \hline
        Score Fusion & HOTA$\uparrow$ & MOTA$\uparrow$ & IDF1$\uparrow$                       & DetA$\uparrow$ & AssA$\uparrow$
                     & HOTA$\uparrow$ & MOTA$\uparrow$ & IDF1$\uparrow$          & DetA$\uparrow$ & AssA$\uparrow$
                     & HOTA$\uparrow$ & MOTA$\uparrow$ & IDF1$\uparrow$         & DetA$\uparrow$ & AssA$\uparrow$ \\ \hline

        \rowcolor{Gray} 
        \ding{51}    & \textbf{69.35} & \textbf{81.04} & \textbf{87.16}                                & \textbf{66.7}7 & \textbf{72.45}
                     & \textbf{68.58} & \textbf{86.07} & \textbf{89.93}          
                     & \textbf{67.65} & \textbf{69.72}
                     & 63.27          & \textbf{92.71} & 66.07                  
                     & \textbf{80.84} & 49.71 \\
                     
        \ding{55}    & 69.25          & 81.01          & 87.09                                         & 66.69          & 72.34
                     & 68.56          & 86.05          & 89.89                  
                     & 67.61          & \textbf{69.72}
                     & \underline{\textbf{63.35}}      & \textbf{92.71}        
                     & \textbf{66.12} & \textbf{80.84} & \textbf{49.83} \\ \hline
        \end{tabular}}
    \label{table:score_fuse}
\end{table*}

Table~\ref{table:score_fuse} assesses the impact of score fusion\cite{bytetrack} in the second association phase of the 2-stage tracker. The baseline configuration with score fusion achieves robust performance with HOTA scores of 69.35, 68.58, and 63.27, respectively, across the datasets. In contrast, omitting the score fusion approach results in marginal performance degradation in MOT17 and MOT20, while bringing slight improvement in DanceTrack. These findings suggest that score fusion provides modest but valuable improvements. Although it shows slight HOTA degradation in DanceTrack, it still maintains detection-related performance. The added stability through score fusion suggests that it can be a beneficial refinement in environments with complex object interactions. Therefore, we keep including the score fusion in our tracker.


\subsection{Post-Processing}

\begin{table*}[t!]
    \centering
    \renewcommand{\arraystretch}{1.25}
    \caption{Comparison of matching with different post-processing methods. AFLink clearly affects the complex scenarios but solely applying gaussian interpolation draws the most balanced performance across datasets.}
    \resizebox{\linewidth}{!}{
        \begin{tabular}{c|c|ccccc|ccccc|ccccc}
        \hline
        \multicolumn{2}{c|}{}  
        & \multicolumn{5}{c|}{MOT17} 
        & \multicolumn{5}{c|}{MOT20} 
        & \multicolumn{5}{c}{DanceTrack} \\ \hline
        AFLink      & Interpolation     & HOTA$\uparrow$ & MOTA$\uparrow$                                                 & IDF1$\uparrow$ & DetA$\uparrow$                                                 & AssA$\uparrow$
                                        & HOTA$\uparrow$ & MOTA$\uparrow$                 & IDF1$\uparrow$ & DetA$\uparrow$                 & AssA$\uparrow$
                                        & HOTA$\uparrow$ & MOTA$\uparrow$                 & IDF1$\uparrow$ & DetA$\uparrow$                 & AssA$\uparrow$  \\ \hline
                               
        \ding{55}   & \ding{55}  
        & 69.35                         & 81.04          & 87.16          
                                        & 66.77          & 72.45
        & 68.58                         & 86.07          & 89.93          
                                        & 67.65          & 69.72
        & 63.27                         & 92.71          & 66.07                                                          & \textbf{80.84} & 49.71 \\

        \ding{51}   & \ding{55}  
        & 68.97                         & 81.04          & 86.28          
                                        & 66.70          & 71.75
        & 68.53                         & 86.04          & 89.72                                                          & 67.61          & 69.65
        & \underline{\textbf{63.68}}    & \textbf{92.72} & \textbf{66.78} 
                                        & 80.82          & \textbf{50.37} \\
        \hline
        \ding{55}   & LI
        & \underline{70.16}             & 82.34          & 87.76                                                          & 67.81          & 73.03
        & \underline{68.91}             & 86.45          & 90.14                                                          & 67.91          & 70.11
        & - & - & - & - & - \\
                
        \rowcolor{Gray}
        \ding{55}   & GSI  
        & \underline{70.34}             & 82.61          & 87.85          
                                        & 67.97          & 73.34
        & \underline{\textbf{69.69}}    & \textbf{86.82} & \textbf{90.19}   
                                        & \textbf{68.66} & \textbf{71.00}
        & - & - & - & - & - \\

        \rowcolor{Gray}
        \ding{55}   & GBI
        & \underline{\textbf{70.79}}    & \textbf{82.78} & \textbf{87.93} 
                                        & \textbf{68.49} & \textbf{73.66}
        & \underline{69.21}             & 86.79          & 90.18 
                                        & 68.25          & 70.39
        & - & - & - & - & - \\ \hline

        \ding{51}   & LI                
        & \underline{69.72}             & 82.32          & 86.86                                                          & 67.72          & 72.21
        & \underline{68.85}             & 86.54          & 89.90                                                          & 67.94          & 69.96
        & - & - & - & - & - \\
                               
        \ding{51}   & GSI  
        & \underline{69.82}             & 82.28          & 86.88                                                          & 67.73          & 72.54
        & \underline{69.62}             & 86.70          & 90.05                                                          & 68.63          & 70.90
        & - & - & - & - & - \\
                               
        \ding{51}   & GBI
        & \underline{70.28}             & 82.64          & 86.99                                                          & 68.32          & 72.80
        & \underline{69.14}             & 86.67          & 90.02                                                          & 68.22          & 70.29
        & - & - & - & - & - \\ \hline
        \end{tabular}}
    \label{table:postprocessing}
\end{table*}

Table~\ref{table:postprocessing} compares several post-processing strategies, including Linear Interpolation (LI)\cite{bytetrack}, Gaussian Score Interpolation (GSI)\cite{gsi, strong_sort}, Gaussian Blur Interpolation (GBI)\cite{gbi, boosttrack}, and Affinity Link (AFLink)\cite{strong_sort}, across MOT17, MOT20, and DanceTrack. Interpolation-based methods are not evaluated on DanceTrack because the dataset does not provide ground-truth annotations for fully occluded objects. Interpolation may link track segments across long occlusions where objects are not visible, leading to erroneous trajectory reconstruction and potential false positives because of the absence of ground-truth annotations. In this manner, we only evaluate Interpolation-based methods on MOT17 and MOT20.

Applying interpolation generally improves performance on MOT17 and MOT20 compared to the baseline without post-processing. In particular, GBI achieves the best performance on MOT17 with HOTA of 70.79, while GSI shows the strongest results on MOT20, achieving the highest HOTA of 69.69 and improving both detection and association metrics. LI also consistently improves the baseline, indicating that reconnecting short-term track fragmentation effectively enhances trajectory continuity.

AFLink shows different behaviors across datasets. While it slightly degrades performance on MOT17 and MOT20, it improves performance on DanceTrack, achieving the best HOTA of 63.68 and the highest association accuracy. This indicates that AFLink is particularly beneficial in highly dynamic scenarios where appearance and motion patterns vary significantly. However, combining AFLink with interpolation methods does not yield additional gains and generally performs slightly worse than using interpolation alone. Overall, Gaussian-based interpolation methods provide the most balanced performance across datasets.


\subsection{Final Baseline}
\begin{table*}[t!]
    \centering
    \renewcommand{\arraystretch}{1.25}
    \caption{Performance comparison between the original SORT baseline and the proposed final baseline. Our tracker integrates the most effective components identified in the ablation studies and shows consistent improvements across MOT17, MOT20, and DanceTrack. Additional gains are obtained when post-processing is applied.}
    \resizebox{\linewidth}{!}{
        \begin{tabular}{c|ccccc|ccccc|ccccc}
        \hline 
        & \multicolumn{5}{c|}{MOT17} 
        & \multicolumn{5}{c|}{MOT20} 
        & \multicolumn{5}{c}{DanceTrack} \\ \hline
        Tracker
        & HOTA$\uparrow$ & MOTA$\uparrow$ & IDF1$\uparrow$ 
        & DetA$\uparrow$ & AssA$\uparrow$
        & HOTA$\uparrow$ & MOTA$\uparrow$ & IDF1$\uparrow$ 
        & DetA$\uparrow$ & AssA$\uparrow$
        & HOTA$\uparrow$ & MOTA$\uparrow$ & IDF1$\uparrow$ 
        & DetA$\uparrow$ & AssA$\uparrow$  \\ \hline
                               
        First baseline(SORT)  
        & 63.91 & 78.66  & 77.85  & 65.21  & 63.17  
        & 66.40 & 84.97  & 86.20  & 66.80  & 66.20  
        & 53.26 & 90.86  & 52.55  & 79.70  & 35.76 \\
        \rowcolor{Gray}
        \textbf{Ours} 
        & 69.35 & 81.04 & 87.16 & 66.77 & 72.45
        & 68.58 & 86.07 & 89.93 & 67.65 & 69.72
        & 63.27 & 92.71 & 66.07 & 80.84 & 49.71 \\ 
        \rowcolor{Gray}
        \textbf{Ours w/ post-processing} 
        & 69.82 & 82.28 & 86.88 & 67.73 & 72.54
        & 69.62 & 86.70 & 90.05 & 68.63 & 70.90
        & - & - & - & - & - \\
        \hline
    \end{tabular}}
\end{table*}

Based on the findings from the component-wise analysis, we construct a final baseline tracker by integrating the most effective design choices. Starting from the basic tracking pipeline, SORT, we adopt a Kalman state vector defined as $(cx, cy, w, h)$\cite{bot_sort} to provide a more direct spatial representation of bounding boxes. To improve robustness in practical scenarios, we incorporate Occlusion-aware Initialization (OAI)\cite{improved}, Confidence Weighted Kalman-Update (CWKU)\cite{conftrack}, and Global Motion Compensation (GMC)\cite{gmc}, which address occlusion handling, improve the reliability of state updates, and compensate for camera-induced motion, respectively.
To further enhance performance, several implementation refinements are applied, including setting the track activation frame count to 2, introducing Detection-Track Confidence Similarity Boost\cite{boosttrack}, and applying an Exponential Moving Average (EMA)\cite{ema} for smoother appearance feature updates and more stable association.
Finally, we employ weighted sum score fusion and a two-stage association strategy\cite{bytetrack}, enabling more reliable matching under varying detection confidence levels. Trajectory continuity is additionally improved through post-processing, Gaussian-based interpolation\cite{gsi,gbi}, establishing a strong and generalized baseline across diverse MOT scenarios.
As shown in Table~\ref{table:postprocessing}, the proposed baseline significantly improves performance over the original SORT tracker across all datasets, demonstrating substantial gains in both detection and association quality. Our final performance improves HOTA of 5.44, 2.18, and 10.01 on MOT18, MOT20, and DanceTrack, respectively. The increase on DanceTrack was the largest, which indicates our new baseline can deal better with complicated scenarios. Similar improvements are observed in IDF1 and AssA, indicating more reliable identity preservation and trajectory consistency.


\section{Conclusion}
\label{sec:con}

In this work, we present a systematic analysis of the tracking-by-detection (TBD) paradigm by examining how individual design components influence the performance of multi-object tracking (MOT) systems. Through controlled experiments across three diverse benchmarks—MOT17, MOT20, and DanceTrack—we investigate the role of key tracking modules, including motion modeling, initialization strategies, data association mechanisms, and post-processing techniques. 

Our study highlights that the effectiveness of a tracker does not arise from a single component but rather from the careful integration of complementary design choices within the TBD pipeline. By evaluating these components under a consistent experimental protocol, we identify combinations that lead to stable and well-balanced tracking performance across datasets with different characteristics, such as dense crowds and highly dynamic motions.

Beyond establishing a strong baseline, our analysis provides practical insights into the design of reliable TBD trackers and offers a clearer understanding of how individual modules interact within the overall tracking framework. We hope that these findings serve as a useful reference for future research and facilitate the development of more robust and versatile MOT systems for real-world applications.





\section*{Declarations}

\noindent\textbf{Funding}
The authors received no specific funding for this work.

\noindent\textbf{Conflict of interest/Competing interests}
The authors declare that they have no competing interests.

\noindent\textbf{Ethics approval and consent to participate}
Not applicable.

\noindent\textbf{Consent for publication}
Not applicable.

\noindent\textbf{Data availability}
Data sharing not applicable to this article as no datasets were generated during the current study. The figures and tables were derived from publicly available literature.

\noindent\textbf{Materials availability}
Not applicable.

\noindent\textbf{Code availability}
The source code supporting the findings of this study will be made publicly available in a GitHub repository upon acceptance of the manuscript.

\noindent\textbf{Author contribution}
Conceptualization, Y.Y and K.S.;
methodology, Y.Y. and K.S.;
investigation, Y.Y and K.S.;
software, Y.Y and K.S.;
formal analysis, Y.Y., K.S., and K.K;
writing--original draft preparation, Y.Y. and K.S;
writing--review and editing, Y.Y., K.S and K.K;
visualization, Y.Y and K.K.;
supervision, C.K.;
project administration, C.K.
All authors have read and approved the final manuscript.

\bibliography{sn-bibliography}

\end{document}